\documentclass[letterpaper]{article} 
\usepackage{aaai2026}  
\usepackage{times}  
\usepackage{helvet}  
\usepackage{courier}  
\usepackage[hyphens]{url}  
\usepackage{graphicx} 
\usepackage{natbib}  
\usepackage{caption} 
\usepackage{algorithm}
\usepackage{algorithmic}

\usepackage{newfloat}
\usepackage{listings}
\DeclareCaptionStyle{ruled}{labelfont=normalfont,labelsep=colon,strut=off} 
\floatstyle{ruled}
\newfloat{listing}{tb}{lst}{}
\floatname{listing}{Listing}
\usepackage{booktabs}
\definecolor{pGray}{RGB}{127, 127, 127}
\definecolor{pRed}{RGB}{192, 0, 0}
\usepackage{amsfonts}
\usepackage{amsmath}
\newtheorem{theorem}{Theorem}[section]
\newtheorem{proof}{Proof}[section]
\usepackage{threeparttable}
\usepackage{adjustbox}
\usepackage{multirow, multicol}
\usepackage{xcolor,colortbl}
\usepackage{titletoc}
\usepackage{subfigure}

\title{Distributional Priors Guided Diffusion for Generating 3D Molecules\\ in Low Data Regimes}

\author {
    Haokai Hong\textsuperscript{\rm 1}, Wanyu Lin\textsuperscript{\rm 1,2,}\thanks{Corresponding author.}, Ming Yang\textsuperscript{\rm 3}, Kay Chen Tan\textsuperscript{\rm 1}
}
\affiliations {
    \textsuperscript{\rm 1}The Department of Data Science and Artificial Intelligence, The Hong Kong Polytechnic University, Hong Kong SAR, China.\\
    \textsuperscript{\rm 2}The Department of Computing, The Hong Kong Polytechnic University, Hong Kong SAR, China.\\
    \textsuperscript{\rm 3}The Department of Applied Physics, The Hong Kong Polytechnic University, Hong Kong SAR, China.\\
    {haokai.hong@connect.polyu.hk},
    {\{wan-yu.lin,kevin.m.yang,kctan\}@polyu.edu.hk}
}

\usepackage{bibentry}

\begin{document}

\maketitle

\begin{abstract}
Can we train a 3D molecule generator using data from dense regions to generate samples in sparse regions? This challenge can be framed as an out-of-distribution (OOD) generation problem. While prior research on OOD generation predominantly targets property shifts, structural shifts---such as differences in molecular scaffolds or functional groups---represent an equally critical source of distributional shifts. This work introduces the Geometric OOD Diffusion Model (\textit{GODD}), a novel diffusion-based framework that enables training on data-abundant molecular distributions while generalizing to data-scarce distributions under distributional structural shifts. Central to our approach is a designated equivariant asymmetric autoencoder to capture distributional structural priors. The asymmetric design allows the model to generalize to unseen structural variations by capturing distributional priors representing distinct distributions. The encoded structural-grained priors guide generation toward sparse regions without requiring explicit training on such data. Evaluated across standard benchmarks encompassing OOD structural shifts (e.g., scaffolds, rings), \textit{GODD} achieves an improvement of 12.6\% in success rate, defined based on molecular validity, uniqueness, and novelty. Furthermore, the framework demonstrates promising performance and generalization on canonical fragment-based drug design tasks, highlighting its utility in learning-based molecular discovery.
\end{abstract}

\section{Introduction}
\label{sec: introduction}
Geometric generative models are proposed to approximate the distribution of complex geometries and are used to generate feature-rich geometries~\cite{RN363, xie2022crystal}. There has been fruitful research progress on 3D molecule generation based on geometric generative modeling. 
Recent representative models for generating 3D molecules \textit{in silico} include autoregressive~\cite{luo2022an}, flow-based models~\cite{NEURIPS2021_21b5680d}, and diffusion models~\cite{pmlr-v162-hoogeboom22a}. Among others, diffusion models have demonstrated their superior performance~\cite{pmlr-v162-hoogeboom22a}. 
However, these generative models require tremendous data to mimic the training distribution. They can barely generate samples that are rare or even absent in the training set, hindering their applicability to {\em de novo} molecule generation~\cite{RN364}.

Taking a canonical molecule dataset -- QM9 as our running example, diverse scaffolds of molecules have varying proportions and frequencies in nature~\cite{QM9, MoleculeNet}. Our initial findings indicate that existing diffusion-based molecular generative models, such as EDM~\cite{pmlr-v162-hoogeboom22a} and GeoLDM~\cite{pmlr-v202-xu23n}, effectively capture the training data distribution, generating molecules with high-frequency scaffolds. However, these models struggle to generate molecules with rare scaffolds (see Table~\ref{tab: preliminary}). With the expressive power of state-of-the-art diffusion-based generators, we ask: {\em Can we train a diffusion model using data from dense regions to generate realistic and valid 3D samples in sparse regions?} 

\begin{table}[t]
\footnotesize
    \centering
    \begin{tabular}{lcccc}
    \toprule
    QM9 & \multicolumn{3}{c}{Scaffold Propotion (\%)}\\
    \midrule
    Domains & In-dist & OOD I & OOD II\\
    \# Molecules & 100,000  & 15,000  & 15,831 \\
    \# Scaffolds & 1,054  & 2,532  & 12,075 \\
    \midrule
    Dataset & 76.4  & 11.5  & 12.1 \\
    \midrule
    EDM   & 91.4  & 2.7   & 4.9  \\
    GeoLDM & 90.6  & 3.5   & 5.9  \\
    \bottomrule
    \end{tabular}%
    \caption{Preliminary results on QM9. In distribution, OOD I and OOD II encompass molecules with high-, low-, and rare-frequency scaffolds, respectively. 
    }
    \label{tab: preliminary}
\end{table}

To address the data scarcity issue, we propose leveraging the concept of {\em out-of-distribution (OOD) generalization} and framing the problem as OOD generation. Our objective, therefore, is to train a generator with data-abundant distribution and steer it to generate samples in sparse regions. The distribution shift generally comes from properties or core structures, such as certain types of scaffolds, ring-structures, fragments, or any sub-structures of molecules~\cite{MoleculeNet, DA-mol-learning}. Certain sets of fragments or properties depict distinct distributions. On the one hand, current fragment-based methodologies focus on in-distribution molecule generation, which is hard to generalize to generate new samples with fragments lying in sparse regions~\cite{unified-guidance-fbdd, linker-design-NMI}. On the other hand, existing works on OOD generation mainly focus on property shifts~\cite{exploring-ood, context-ood}. They usually utilize a naive property predictor for guidance, where the properties are scalars. Due to the sparsity of certain 3D structures, it is imperative to design new OOD generative frameworks to deal with structural shifts.

\begin{figure*}[t]
\begin{center}
\centerline{
    \includegraphics[width=0.95\textwidth]{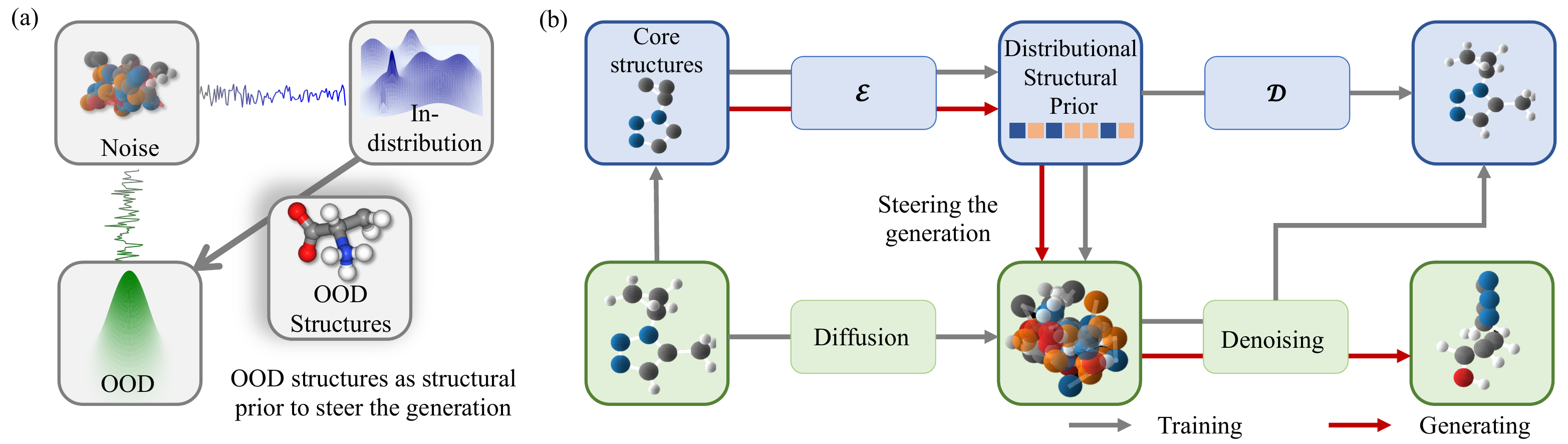}
}
\caption{ The Illustration of the Proposed GODD Framework. (a): GODD leverages OOD structures as priors to guide generation toward data-sparse regions. (b): Training pipeline and generation pipeline for our proposed GODD framework.
}
\label{fig: basic idea}
\end{center}
\end{figure*}

This paper presents a novel Geometric OOD Diffusion Model (\textit{GODD}) that employs distributional structural priors to guide 3D molecule generation in data-sparse regions. To enable OOD generation under structural shifts, \textit{GODD} learns generalizable and equivariant structural representations, termed distributional structural priors, which are integrated into the denoising process. We utilize an asymmetric encoder-decoder architecture, inspired by the success of asymmetric autoencoders in generalizable representation learning, to characterize these priors. This design facilitates transferable learning across distributions, enabling generalization to unseen structural variations, such as OOD scaffolds or ring structures. The \textit{GODD} workflow is illustrated in Figure \ref{fig: basic idea}, with our main contributions outlined below:

\par
{\em First}, to the best of our knowledge, we are the first study to tackle 3D molecule generation in data-sparse regions and frame the problem as an out-of-distribution generation problem under structural shift. 
We ensure and theoretically prove that the structural priors extracted by the designed asymmetric autoencoder are $\mathrm{SE(3)}$-equivariant. Our proposed framework does not require additional training on OOD data.
\par
{\em Second}, we evaluate out-of-distribution generation setting with benchmarking datasets. We compare it with alternative baselines, including vanilla generative models, fragment-based drug design methods, and OOD generative models. Besides, we empirically validate the effectiveness of asymmetric design in OOD generation with ablation studies. Extensive experimental results show that the structural priors enable the model to generate molecules with desired OOD structural variations in data-sparse regions. The success rate of molecules generated by \textit{GODD} is improved by up to 12.6\% compared with existing methods. 
\par
{\em Third}, we demonstrate that our generative framework, guided by structural priors, can be applied to fragment-based OOD generation. We verify that our framework can be readily adapted to link multiple fragments under OOD settings. Specifically, we evaluate our method with a canonical fragment-based drug design task---linker design---and showed that the proposed method exhibits promising performance in fragment linking within the OOD context~\cite{linker-design-NMI}.

\section{Problem Setup}
\label{sec: background}
{\bf Notations:} Let $d$ be the dimensionality of node features; a 3D molecule can be represented as a point cloud denoted as $\mathcal{G} = \langle\mathbf{x},\mathbf{h}\rangle$, where $\mathbf{x} = (\mathbf{x}_1,\dots,\mathbf{x}_N)\in\mathbb{R}^{N\times3}$ is the atom coordinate matrix and $\mathbf{h} = (\mathbf{h}_1,\dots,\mathbf{h}_N)\in\mathbb{R}^{N\times d}$
is the node feature matrix containing atomic type, charge features, etc. For a given molecule $\mathcal{G}$, the core structure~\footnote{We recognize the distinct use of ``structure'' here versus in ``structure-based drug design'' but retain it to denote molecular substructures for clarity and readability in this work.} is a subgraph of the original molecule, represented as $\mathcal{G}^{f}=\langle\mathbf{x}^{f},\mathbf{h}^{f}\rangle$. We explore three important substructures in this work, including scaffold, ring-structure, and fragments. Specifically, the scaffold is its structural framework~\cite{doi:10.1021/jm9602928}, termed as ``chemotypes''. The ring structures are also essential substructures in chemistry and biology~\cite{RN367, Simon_ring_number, RITCHIE20091011}, which could also be a factor that incurs the distribution shift. Fragment, in drug design and biology, refers to a small, low molecular weight compound that binds weakly but specifically to a biological target, serving as a foundational scaffold in fragment-based drug discovery (FBDD)~\cite{fbdd-rise, unified-guidance-fbdd, linker-design-NMI}. 

{\bf Out-of-Distribution (OOD) Generation Problem:} We consider the problem of OOD generation in the following two scenarios: OOD scaffold and OOD ring-structure generation, respectively. Given a collection of molecules as training samples and corresponding in distributional substructure set (including scaffold, ring-structure, or fragments) denoted as $\{\mathcal{G}_{I}\}$, $\{\mathcal{G}^{f}_{I}\}$, respectively. OOD generation aims to learn a generative model that can generate valid and novel molecules falling into a new distribution, where the corresponding structure set is $\{\mathcal{G}^{f}_{O}\}$, and the OOD structure set is unseen during training, a.k.a. $\{\mathcal{G}^{f}_{I}\}\cap \{\mathcal{G}^{f}_{O}\} = \emptyset $. We review OOD generation and fragment-based drug design in Appendix~\ref{sec: related work}.
\section{Method}
\label{sec: method}


\subsection{Equivariant Asymmetric Autoencoder}
\label{sec: method: sec: emae}
\textbf{Distributional Structural Prior.}\ \ For a given substructure $\mathcal{G}^{f}=\langle\mathbf{x}^{f},\mathbf{h}^{f}\rangle$, the distributional structural prior learned from the substructure ($\mathcal{F}$) is defined as $\mathcal{F}=\langle\mathbf{f}_{\mathbf{x}},\mathbf{f}_{\mathbf{h}}\rangle$.
\par
\par
\textbf{Asymmetric Autoencoder.}\ \ The asymmetric autoencoder comprises an encoder $\mathcal{E}$, which maps substructure $\mathcal{G}^{f}$ to a latent space, represented as $\mathbf{f}_{\mathbf{x}}, \mathbf{f}_{\mathbf{h}} =\mathcal{E}(\mathbf{x}^{f}, \mathbf{h}^{f})$. Additionally, it includes a decoder $\mathcal{D}$ that reconstructs the latent representation back to the original molecular space, denoted as $\hat{\mathbf{x}}, \hat{\mathbf{h}} = \mathcal{D}(\mathbf{f}_{\mathbf{x}}, \mathbf{f}_{\mathbf{h}})$. Our autoencoder reconstructs the input by predicting the coordinates and features of complete atoms. The loss function computes the mean squared error (MSE) between the reconstructed and original molecules in the original molecular space. The autoencoder can be trained by minimizing the reconstruction objective, expressed as $\boldsymbol{f}(\mathcal{G}, \mathcal{D}(\mathcal{E}(\mathcal{G}^{f})))$. The encoder of the autoencoder functions solely on the substructure $\mathcal{G}^{f}$, while the decoder reconstructs the input from the latent representation to the complete molecule $\mathcal{G}$. This asymmetric encoder-decoder design offers promising generalization~\cite{He_2022_CVPR} to the latent features. These features serve as structural priors and empower the model to generate molecules with unseen substructures.
\par
\textbf{Equivariant Asymmetric Autoencoder.}\ \ However, naively applying an autoencoder in the geometric domain is non-trivial. The diffusion model within the overall framework operates in 3D molecular space and necessitates conditions to be either equivariant or invariant. Therefore, it is crucial to ensure the equivariance of the conditions extracted by the autoencoder. To achieve this, we design our asymmetric autoencoder based on the Equivariant Graph Neural Networks (EGNNs)~\cite{pmlr-v139-satorras21a}, thereby incorporating equivariance into both the encoder $\mathcal{E}_{\phi}$ and decoder $\mathcal{D}_{\vartheta}$, where $\phi$ and $\vartheta$ are two learnable EGNNs. equivariant design ensures that the latent representations $\mathbf{f}_{\mathbf{x}}$ and $\mathbf{f}_{\mathbf{h}}$ encoded by the encoder from substructures are $3$-D equivariant and $k$-d invariant, respectively. Consequently, Equivariant Asymmetric Autoencoder (EAAE) extracts both invariant and equivariant conditions, as expressed below:
\begin{align}
    \mathbf{R}\mathbf{f}_{\mathbf{x}}+\boldsymbol{t}, \mathbf{f}_{\mathbf{h}} = &\mathcal{E}_{\phi}(\mathbf{R}\mathbf{x}^{f}+\boldsymbol{t}, \mathbf{h}^{f}) \label{eq: encoder-equivariant}\\
    \mathbf{R}\hat{\mathbf{x}}+\boldsymbol{t}, \hat{\mathbf{h}} = &\mathcal{D}_{\vartheta}(\mathbf{R}\mathbf{f}_{\mathbf{x}}+\boldsymbol{t}, \mathbf{f}_{\mathbf{h}}),\label{eq: decoder-equivariant}
\end{align}
for all rotations $\mathbf{R}$ and translations $\mathbf{t}$. Detailed architecture information about the asymmetric autoencoder can be found in Appendix~\ref{appendix: emae}. The point-wise latent space adheres to the inherent structure of geometries $\mathcal{G}^{f}$, which facilitates learning conditions for the diffusion model and results in high-quality molecule design.
\par
Following~\cite{pmlr-v162-hoogeboom22a}, to ensure that linear subspaces with the center of gravity always being zero can induce translation-invariant distributions, we define distributions of substructures $\mathbf{x}^{f}$, structural priors $\mathbf{f}_{x}$, and reconstructed $\hat{\mathbf{x}}$ on the subspace that $\sum_{i}\mathbf{x}^{f}_{i}$ (or $\mathbf{f}_{x, i}$ and $\hat{\mathbf{x}_{i}}$) $=0$. Then the encoding and decoding processes can be formulated by $q_{\phi}(\mathbf{f}_{\mathbf{x}}, \mathbf{f}_{\mathbf{h}}|\mathbf{x}^{f},\mathbf{h}^{f})=\mathcal{N}(\mathcal{E}_{\phi}(\mathbf{x}^{f},\mathbf{h}^{f}), \sigma_{0}\boldsymbol{I})$ and $p_{\vartheta}(\mathbf{x},\mathbf{h}|\mathbf{f}_{\mathbf{x}},\mathbf{f}_{\mathbf{h}})=\prod_{i=1}^{N}p_{\vartheta}(x_i,h_i|\mathbf{f}_{\mathbf{x}},\mathbf{f}_{\mathbf{h}})$ and the \textit{EAAE} can be optimized by:
\begin{equation}
\begin{aligned}
\label{eq: emae loss}
    \mathcal{L}_{\textrm{\textit{EAAE}}}(\mathcal{G},\mathcal{G}^{f})
    = \mathbb{E}_{q_{\phi}(\mathbf{f}_{\mathbf{x}}, \mathbf{f}_{\mathbf{h}}|\mathbf{x}^{f},\mathbf{h}^{f})}p_{\vartheta}(\mathbf{x},\mathbf{h}|\mathbf{f}_{\mathbf{x}},\mathbf{f}_{\mathbf{h}})\\
    -\mathrm{KL} [q_{\phi}(\mathbf{f}_{\mathbf{x}}, \mathbf{f}_{\mathbf{h}}|\mathbf{x}^{f},\mathbf{h}^{f})||\prod_{i}^{N}\mathcal{N}({f}_{\mathbf{x},i},{f}_{\mathbf{h}, i}|0,\mathbf{I})],
\end{aligned}
\end{equation}
where $\mathbb{E}_{q_{\phi}(\mathbf{f}_{\mathbf{x}}, \mathbf{f}_{\mathbf{h}}|\mathbf{x}^{f},\mathbf{h}^{f})}p_{\vartheta}(\mathbf{x},\mathbf{h}|\mathbf{f}_{\mathbf{x}},\mathbf{f}_{\mathbf{h}})$ is the asymmetric reconstruction loss and is calculated as $L_2$ norm or cross-entropy for continuous or discrete features. $\textrm{KL}[q_{\phi}(\mathbf{f}_{\mathbf{x}}, \mathbf{f}_{\mathbf{h}}|\mathbf{x}^{f},\mathbf{h}^{f})||\prod_{i}^{N}\mathcal{N}({f}_{x},{f}_{h}|0,\mathbf{I}])$ is a regularization term between $q_{\phi}$ and standard Gaussians. $\mathcal{L}_{\textrm{\textit{EAAE}}}$ is the standard VAE loss and is the variational lower bound of log-likelihood. The equivariance of the loss, which is crucial for geometric graph generation, is expressed as follows:
\begin{theorem}
\label{thm: loss emae}
If $\mathcal{L}_{\textrm{\textit{EAAE}}}$ is an $SE(3)$-invariant variational lower bound to the log-likelihood, i.e., for any substructure $\langle\mathbf{x}^{f}, \mathbf{h}^{f}\rangle$ and molecule $\langle\mathbf{x}, \mathbf{h}\rangle$, we have $\forall~\mathbf{R}~\textrm{and}~\mathbf{t}, ~\mathcal{L}_{\textrm{\textit{EAAE}}}(\mathbf{x}, \mathbf{h}, \mathbf{x}^{f}, \mathbf{h}^{f})=\mathcal{L}_{\textrm{\textit{EAAE}}}(\mathbf{R}\mathbf{x}+\mathbf{t}, \mathbf{h}, \mathbf{R}\mathbf{x}^{f}+\mathbf{t}, \mathbf{h}^{f})$.
\end{theorem}

Theorem~\ref{thm: loss emae} guarantees that the asymmetric autoencoder is $\mathrm{SE}(3)$-equivariant, ensuring that the extracted conditioning features obey the required symmetry constraints and that the conditional denoising step of the geometric diffusion model remains equivariant. The full proof is provided in Appendix~\ref{appendix: emae loss}. In summary, \textit{EAAE} encodes the structural prior $\mathcal{G}^{f}$ with the encoder $\mathcal{E}$ to produce equivariant latent features $\mathbf{f}_{\mathbf{x}}$ and invariant latent features $\mathbf{f}_{\mathbf{h}}$. These features serve two roles: they are fed to the decoder $\mathcal{D}$ for reconstruction, which regularizes the latent space, and they are used as symmetry-aware conditions to guide the diffusion denoising process. The precise conditioning mechanism is detailed in the next section.
\subsection{Structural Prior Steered Diffusion Model}
\label{sec: method: sec: scdm}
\par
Generally, geometric diffusion models are capable of controllable generation with given conditions $s$ by modeling conditional distributions $p(\mathbf{z}|s)$. This modeling in DMs can be implemented with conditional denoising networks $\epsilon_{\theta}(\mathbf{z}, t, s)$ with the critical difference that it takes additional inputs $s$. However, an underlying constraint of such use is the assumption that $s$ is invariant. By contrast, a fundamental challenge for our method is that the conditions for the DM contain not only invariant features $\mathbf{f}_{\mathbf{h}}$ but also equivariant features $\mathbf{f}_{\mathbf{x}}$. This requires the distribution $p_{\theta}(\mathbf{z}_{0:T})$ of our DMs to satisfy the critical invariance:
\begin{equation}
\label{eq: diffusion equivariant}
\begin{aligned}
    \forall~\mathbf{R},~&p_{\theta}(\mathbf{z}_{\mathbf{x}}, \mathbf{z}_{\mathbf{h}}, \mathbf{f}_{\mathbf{x}},\mathbf{f}_{\mathbf{h}})=p_{\theta}(\mathbf{R}\mathbf{z}_{\mathbf{x}}, \mathbf{z}_{\mathbf{h}}, \mathbf{R}\mathbf{f}_{\mathbf{x}},\mathbf{f}_{\mathbf{h}}),
\end{aligned}
\end{equation}
where $\mathbf{z}_{\mathbf{x}}$ and $ \mathbf{z}_{\mathbf{h}}$ are the noises. To achieve this, we should ensure that (1) the initial distribution $p(\mathbf{z}_{\mathbf{x},T}, \mathbf{z}_{\mathbf{h},T},\mathbf{f}_{\mathbf{x}},\mathbf{f}_{\mathbf{h}})$ is invariant, which is already satisfied since $\mathbf{z}_{\mathbf{x},T}$ is projected down by subtracting its center of gravity after sampling from standard Gaussian noise. With the $\mathbf{f}_{\mathbf{x}},\mathbf{f}_{\mathbf{h}}$ is obtained by equivariant $\mathcal{E}_{\phi}$ (Equations~\ref{eq: encoder-equivariant}); (2) the conditional reverse processes via $\theta$, which is expressed as $p_{\theta}(\mathbf{z}_{\mathbf{x},t-1}, \mathbf{z}_{\mathbf{h},t-1}|\mathbf{z}_{\mathbf{x},t}, \mathbf{z}_{\mathbf{h},t},\mathbf{f}_{\mathbf{x}},\mathbf{f}_{\mathbf{h}})$, are equivariant:
\begin{equation}
    \begin{aligned}
        \forall~\mathbf{R},~ p_{\theta}(\mathbf{z}_{\mathbf{x},t-1}, \mathbf{z}_{\mathbf{h},t-1}|\mathbf{z}_{\mathbf{x},t}, \mathbf{z}_{\mathbf{h},t},\mathbf{f}_{\mathbf{x}},\mathbf{f}_{\mathbf{h}}) \\
        =  p_{\theta}(\mathbf{R}\mathbf{z}_{\mathbf{x},t-1}, \mathbf{z}_{\mathbf{h},t-1},|\mathbf{R}\mathbf{z}_{\mathbf{x},t}, \mathbf{z}_{\mathbf{h},t},\mathbf{R}\mathbf{f}_{\mathbf{x}},\mathbf{f}_{\mathbf{h}}),
    \end{aligned}
\end{equation}
this can be realized by implementing the denoising network $\boldsymbol{\epsilon}_{\theta}$ with EGNN that satisfy the following equivariance:
\begin{equation}
    \begin{aligned}
        \forall~\mathbf{R}~\textrm{and}~\mathbf{t},~
        \mathbf{R}\mathbf{z}_{\mathbf{x},t-1}+\mathbf{t},\mathbf{z}_{\mathbf{h},t-1}= \\
        \boldsymbol{\epsilon}_{\theta}(\mathbf{R}\mathbf{z}_{\mathbf{x},t}+\mathbf{t},\mathbf{z}_{\mathbf{h},t},\mathbf{R}\mathbf{f}_{\mathbf{x}}+\mathbf{t},\mathbf{f}_{\mathbf{h}},t).
    \end{aligned}
\end{equation}
To keep translation invariance, all the intermediate states $\mathbf{z}_{\mathbf{x},t}, \mathbf{z}_{\mathbf{h},t}$ are also required to lie on the subspace by $\sum_{i}\mathbf{z}_{\mathbf{x},t,i}=0$ by moving the center of gravity. Analogous to the equation in training diffusion model~\cite{NEURIPS2020_4c5bcfec}, now we can train the Distributional Prior Steered Diffusion Model (DSDM) by minimizing the loss:
\begin{equation}
\begin{aligned}
\label{eq: scdm loss}
    &\mathcal{L}_{\textrm{\textit{DSDM}}}(\mathcal{G},\mathcal{G}^{f})=\\
    &\mathbb{E}_{\mathcal{G},\mathcal{E}(\mathcal{G}^{f}),\epsilon,t}\left[
    w(t)\Vert
    \boldsymbol{\epsilon}-\boldsymbol{\epsilon}_{\theta}(\mathbf{z}_{\mathbf{x},t}, \mathbf{z}_{\mathbf{h},t},\mathbf{f}_{\mathbf{x}},\mathbf{f}_{\mathbf{h}},t)
    \Vert^{2}
    \right],
\end{aligned}
\end{equation}
with $w(t)$ simply set as $1$ for all steps $t$, where $w(t)=\frac{\beta_{t}}{2\sigma_{t}^{2}\alpha_{t}(1-\bar\alpha_{t})}$ is the reweighting term~\cite{NEURIPS2020_4c5bcfec}. As the EGNN only receives atomic coordinates and features $\mathbf{z}_{\mathbf{x},t}$ and $\mathbf{z}_{\mathbf{h},t}$, we concatenate $\mathbf{f}_{\mathbf{x}}$ and $\mathbf{f}_{\mathbf{h}}$ to the node features $\mathbf{z}_{\mathbf{h},t}$. Specifically, with node features $\mathbf{z}_{\mathbf{h},t} \in \mathbb{R}^{N \times d}$, a time-step embedding $\mathbf{t}\in\mathbb{R}^{N \times 1}$, $\mathbf{f}_{\mathbf{x}} \in \mathbb{R}^{N' \times 3}$, and $\mathbf{f}_{\mathbf{h}} \in \mathbb{R}^{N' \times k}$, the EGNN within the denoising network $\boldsymbol{\epsilon}_{\theta}$ processes coordinates $\mathbf{z}_{\mathbf{x},t} \in \mathbb{R}^{N \times 3}$ and concatenated features $\mathbf{z}_{\mathbf{h},t} \in \mathbb{R}^{N \times (d+3+k+1)}$. Since the number of atoms of the substructure ($N'$) is less than the number of atoms of the molecule ($N$), zeros are padded to $\mathbf{f}_{\mathbf{x}}$ and $\mathbf{f}_{\mathbf{h}}$. We briefly introduced diffusion models in Appendix~\ref{appendix: dm detail}.
\par
\subsection{Training and Generating OOD Samples}
\label{sec: method: sec: damg}
\textbf{Training.}~The training loss of the entire framework can be formulated as $\mathcal{L}=\mathcal{L}_{\textrm{\textit{EAAE}}}+\mathcal{L}_{\textrm{\textit{DSDM}}}$. To make the training loss tractable, we also show that $\mathcal{L}$ is theoretically an SE(3)-invariant variational lower bound of the log-likelihood, and we can have:
\begin{theorem}
\label{thm: loss SE3}
Let $\mathcal{L} := \mathcal{L}_{\textrm{\textit{EAAE}}} + \mathcal{L}_{\textrm{\textit{DSDM}}}$. With certain weights $w(t)$, $\mathcal{L}$ is an $SE(3)$-invariant variational lower bound to the log-likelihood. 
\end{theorem}
 
Given the above training loss and Theorem~\ref{thm: loss SE3}, we can optimize \textit{GODD} via back-propagation with the reparameterizing trick~\cite{kingma2013auto}. We provide the detailed proof of Theorem~\ref{thm: loss SE3} in Appendix~\ref{appendix: proof}, and a formal description of the optimization procedure in Algorithm~\ref{alg: training} in Appendix~\ref{appendix: algorithm}. We follow the process of EDM~\cite{pmlr-v162-hoogeboom22a} regarding the representation for continuous features $\mathbf{x}$ and categorical features $\mathbf{h}$. For clarity, we provide the details in Appendix~\ref{appendix: MMF}.

\par
\textbf{Generating OOD Molecules.} With \textit{GODD} trained on dataset $\{\mathcal{G}_{I}\}$ and given an OOD scaffold/ring-structure $\mathcal{G}_{O}^{f}$, we can perform OOD molecule generation (a scaffold OOD generative process is illustrated in Figure~\ref{fig: scaffold conditional design} in Appendix~\ref{appendix: visualization}). To sample from the model, one first inputs the $\mathcal{G}_{O}^{f}$ into the encoder $\mathcal{E}_{\phi}$ and obtains the latent representation of $\mathcal{G}_{O}^{f}$ denoted as structural prior $\langle\mathbf{f}_{\mathbf{x}},\mathbf{f}_{\mathbf{h}}\rangle$ via reparameterization. With the OOD structural prior as condition, the framework first samples $\mathbf{z}_{\mathbf{x},T},\mathbf{z}_{\mathbf{h},T}\sim\mathcal{N}_{x,h}(\mathbf{0},\mathbf{I})$ and then iteratively samples  $\mathbf{z}_{\mathbf{x},t-1},\mathbf{z}_{\mathbf{h},t-1}\sim p_{\theta}(\mathbf{z}_{\mathbf{x},t-1},\mathbf{z}_{\mathbf{h},t-1}|\mathbf{z}_{\mathbf{x},t},\mathbf{z}_{\mathbf{h},t},\mathbf{f}_{\mathbf{x}},\mathbf{f}_{\mathbf{h}})$. Finally, the output molecule represented as $\langle\mathbf{x},\mathbf{h}\rangle$ is sampled from $p(\mathbf{z}_{\mathbf{x},0},\mathbf{z}_{\mathbf{h},0}|\mathbf{z}_{\mathbf{x},1},\mathbf{z}_{\mathbf{h},1},\mathbf{f}_{\mathbf{x}},\mathbf{f}_{\mathbf{h}})$. The pseudo-code of the OOD generation is provided in Algorithm~\ref{alg: sampling} in Appendix~\ref{appendix: algorithm}.
\section{Experiments}
\label{sec: experiments}
\renewcommand\arraystretch{0.8}
\subsection{Experiment Setup}
\textbf{Datasets and Tasks.} We evaluate over QM9~\cite{QM9} and the GEOM-DRUG~\cite{geomdrug}. Specifically, QM9 is a standard dataset that contains molecular properties and atom coordinates for 130,000 3D molecules with up to 9 heavy atoms and up to 29 atoms. GEOM-DRUG encompasses around 450,000 molecules, each with an average of 44 atoms and a maximum of 181. Dataset details and experimental parameters are presented in Appendices \ref{appendix: qm9}, \ref{appendix: drug}, and \ref{appendix: parameters}.

\textit{Ring-Structure Molecule Generation.} Ring-structure variations cause distribution shifts in this task. Using RDKit~\cite{landrum2016rdkit}, we classified QM9 dataset molecules into nine groups (0–8 rings), with molecule counts decreasing as ring numbers increase. The QM9 dataset was split into training (0–3 rings) and five target distributions (4–8 rings). Figure~\ref{fig: ring qm9} in the Appendix shows example molecules with 0–8 rings. For the GEOM-DRUG dataset, molecules have 0–14 or 22 rings. The training set includes 0–10 rings, with five target distributions (11–14 and 22 rings), each containing fewer than 100 molecules, indicating data-sparse regions.

\textit{Scaffold Molecule Generation.} Scaffold variations in this task cause distribution shifts. Using RDKit~\cite{landrum2016rdkit}, we analyzed the scaffolds of molecules in the QM9 dataset, marking those without scaffolds as `-' and including them in the total scaffold count. The dataset was split based on scaffold frequency: the in-distribution set comprised 100,000 molecules and 1,054 scaffolds, most appearing $\geq$ 100 times; out-of-distribution I (OOD I) included 15,000 molecules and 2,532 scaffolds, most appearing 10–100 times; and out-of-distribution II (OOD II) contained 15,831 molecules and 12,075 scaffolds, each appearing $<$ 10 times. We aim to train a generative model on the in-distribution data to produce effective molecules for target distributions, such as OOD I and II.

\textit{Linker Design.} We evaluated the framework on the linker design task, demonstrating \textit{GODD}'s proof-of-concept in canonical fragment-based design under out-of-distribution (OOD) settings. Notably, the GEOM-LINKER dataset exhibits fragment shifts driven by molecular ring counts, with molecules having more than eight rings being highly sparse. For evaluation, we divided the GEOM-LINKER dataset by ring count, designating molecules with sparse ring numbers as the OOD test set. Additional details on the GEOM-LINKER dataset and related work are provided in Appendices~\ref{appendix: linker} and \ref{sec: related work}. 



\textbf{Baselines.} We evaluate four methodological categories for comprehensive performance comparisons: unconditional generation, conditional generation, OOD generation, and fragment-based drug design frameworks. 1. \textbf{Unconditional Generation}: We benchmark four state-of-the-art 3D unconditional molecule diffusion models—EDM~\cite{pmlr-v162-hoogeboom22a}, GeoLDM~\cite{pmlr-v202-xu23n}, EquiFM~\cite{equifm}, and GeoBFN~\cite{geobfn}—to assess the OOD generation capabilities of our proposed GODD. 2. \textbf{Conditional Generation}: We adapt EEGSDE~\cite{bao2023equivariant} and modify EDM and GeoLDM (denoted C-EDM, C-GeoLDM, and EEGSDE) for scalar-based conditional generation using ring counts as the conditional feature to evaluate GODD's performance in OOD ring-structure generation. 3. \textbf{OOD-Specific Frameworks}: We compare GODD against MOOD~\cite{exploring-ood} and CGD~\cite{context-ood}, two dedicated OOD generative frameworks, for ring-structure molecule generation tasks. 4. \textbf{Fragment-Based Methods}: We evaluate DiffLinker~\cite{linker-design-NMI} and LinkerNet~\cite{linkernet-nips} as fragment-based baselines, using the same scaffold/ring-structure/fragment inputs as GODD. DiffLinker employs a diffusion model for multi-fragment linking, while LinkerNet uses Riemannian manifold-based diffusion for enhanced geometric fidelity.

\par
\textbf{Metrics.} Our objective is to generate effective 3D molecules in data-sparse regions. A generated sample is effective only when it falls into the target distribution while it is valid, unique, and novel simultaneously. Therefore, our evaluation metrics can be defined as follows:

1. \textbf{\textit{Proportion (P)}}: Given an OOD scaffold/ring set $\{\mathcal{G}^{f}_{O}\}$, proportion describes the percentage of molecules that contain the desired scaffold/ring-structure in $\{\mathcal{G}^{f}_{O}\}$ among generated valid samples;~
2. \textit{\textbf{Coverage (C)}}: Coverage describes the percentage of scaffold set of the generated samples (denoted as $\{\mathcal{G}^{f}_{G}\}$) in the OOD scaffold set $\{\mathcal{G}^{f}_{O}\}$, which is expressed as $C=|\{\mathcal{G}^{f}_{G}\}|/|\{\mathcal{G}^{f}_{O}\}|$;~
3. \textit{\textbf{Target atom stability (AS)}}: The ratio of atoms that has the correct valency with the desired scaffold/ring-structure among all generated molecules;~
4. \textit{\textbf{Target molecule stability (MS)}}: The ratio of generated molecules contains the desired scaffold/ring-structure, and all atoms are stable. GEOM-DRUG dataset has nearly 0\% molecule-level stability, so this metric is generally ignored on GEOM-DRUG~\cite{pmlr-v162-hoogeboom22a};~
5. \textit{\textbf{Target validity (V)}}: The percentage of valid molecules among all the desired molecules, which is measured by RDkit~\cite{landrum2016rdkit} and widely used for calculating validity~\cite{pmlr-v162-hoogeboom22a, pmlr-v202-xu23n});~
6. \textit{\textbf{Target novelty (N)}}: The percentage of novel molecules among all the desired valid molecules, the novel molecule is different from training samples;~
7. \textit{\textbf{Success rate (S)}}: The ratio of generated valid, unique, and novel molecules that contain the desired scaffold/ring-structure.

\subsection{Results and Analysis}
\begin{table*}[t]
\small
  \centering
  
  \begin{threeparttable}
    \begin{tabular}{c|cccc||ccccc|ccccc}
    \toprule
    Metrics $\uparrow$ & \multicolumn{4}{c||}{P (\%) in Distributions} & \multicolumn{5}{c|}{P (\%) in OOD Generation} & AS    & MS    & V     & N     & S \\
    \midrule
    \# Rings & 0     & 1     & 2     & 3     & 4     & 5     & 6     & 7     & 8     & \multicolumn{5}{c}{Averaged Metrics (\%)} \\
    \midrule
    QM9   & 10.2  & 39.3  & 27.6  & 15.1  & 4.4   & 2.7   & 0.6   & 0.2   & 0.0   & 99.0  & 95.2  & 97.7  & -     & - \\
    \midrule
    EDM   & 10.5  & 39.8  & 28.0  & 14.5  & 4.0   & 2.9   & 0.2   & 0.1   & \multicolumn{1}{c|}{0.0} & 11.0  & 9.7   & 10.5  & 6.8   & 6.3 \\
    GeoLDM & 12.0  & 38.6  & 27.0  & 15.3  & 4.6   & 2.2   & 0.2   & 0.1   & \multicolumn{1}{c|}{0.0} & 10.9  & 9.1   & 10.4  & 6.4   & 5.9 \\
    EquiFM & 12.1  & 44.1  & 29.8  & 11.8  & 1.7   & 0.5   & 0.0   & 0.0   & \multicolumn{1}{c|}{0.0} & 11.0  & 9.7   & 10.4  & 6.8   & 6.3 \\
    GeoBFN & 2.8   & 41.5  & 32.1  & 15.7  & 4.7   & 2.7   & 0.3   & 0.1   & \multicolumn{1}{c|}{0.0} & 10.9  & 9.1   & 10.4  & 6.7   & 6.1 \\
    \midrule
    C-EDM & 98.9  & 94.2  & 80.8  & 64.4  & 12.6  & 26.8  & 0.3   & 0.1   & 0.0   & 41.3  & 33.9  & 38.0  & 27.3  & 24.1 \\
    C-GeoLDM & 97.1  & 89.4  & 74.2  & 52.4  & 22.3  & 22.7  & 0.9   & 0.2   & 0.0   & 39.1  & 31.5  & 35.7  & 28.3  & 25.0 \\
    EEGSDE & 98.4  & 92.2  & 77.6  & 58.2  & 14.1  & 17.6  & 0.3   & 0.0   & 0.0   & 39.1  & 31.1  & 35.7  & 27.2  & 24.2 \\
    \midrule
    MOOD  & 80.7  & 87.1  & 86.1  & 73.3  & 34.1  & 32.3  & 10.3  & 0.2   & \multicolumn{1}{c|}{0.0} & 44.3  & 39.0  & 42.1  & 27.7  & 25.5 \\
    CGD   & 82.3  & 84.8  & 86.2  & 83.6  & 34.4  & 22.4  & 10.3  & 10.1  & \multicolumn{1}{c|}{0.0} & 45.5  & 40.0  & 43.2  & 28.4  & 26.2 \\
    \midrule
    DiffLinker & 99.7  & 99.9  & 99.0  & 91.4  & 84.7  & 75.6  & 74.6  & 63.2  & 59.4  & 78.8  & 53.7  & 70.3  & 48.4  & 26.4 \\
    LinkerNet & 99.8  & 99.6  & 88.8  & 87.2  & 83.2  & 73.7  & 66.1  & 64.7  & 59.2  & 76.0  & 51.6  & 72.2  & 54.4  & 37.0 \\
    \midrule
    \rowcolor{lightgray}
    \textit{GODD} & \textbf{99.9} & \textbf{99.8} & \textbf{99.1} & \textbf{97.6} & \textbf{92.5} & \textbf{89.7} & \textbf{78.7} & \textbf{88.2} & \textbf{82.1} & \textbf{83.1} & \textbf{54.0} & \textbf{77.9} & \textbf{70.3} & \textbf{40.5} \\
    \bottomrule
    \end{tabular}%
    \begin{tablenotes}
    \small
        \item[] \textbf{C-}: C-EDM and C-GeoLDM are trained with conditioning on ring counts.
    \end{tablenotes}
    \end{threeparttable}
    \caption{Results of molecule proportion in terms of ring-number (P), atom stability (AS), molecule stability (MS), validity (V), novelty (N), and success rate (S). QM9 contains 36 eight-ring molecules, and the proportion is nearly 0.}
  \label{tab: ring coverage}%
\end{table*}%
\par
\begin{table}[t]
    \centering
    \begin{threeparttable}
    \begin{tabular}{c|ccccc}
    \toprule
    \multicolumn{6}{c}{Averaged metrics (\%) $\uparrow$} \\
    \midrule
    Method & P     & AS    & V     & N     & S \\
    \midrule
    GEOM-DRUG & 0.00  & 86.50 & 99.90 & -     & - \\
    \midrule
    EDM   & 0.00  & 0.00  & 0.00  & 0.00  & 0.00 \\
    GeoLDM & 0.00  & 0.00  & 0.00  & 0.00  & 0.00 \\
    EquiFM & 0.00  & 0.00  & 0.00  & 0.00  & 0.00 \\
    GeoBFN & 0.00  & 0.00  & 0.00  & 0.00  & 0.00 \\
    \midrule
    MOOD  & 0.00  & 0.00  & 0.00  & 0.00  & 0.00 \\
    CGD   & 0.00  & 0.00  & 0.00  & 0.00  & 0.00 \\
    \midrule
    DiffLinker & 6.29  & 5.21  & 5.01  & 6.25  & 4.16 \\
    LinkerNet  & 10.15 & 8.36  & 7.92  & 9.73  & 7.75 \\
    \midrule
    \rowcolor{lightgray}
    \textit{GODD} & \textbf{13.8} & \textbf{11.4} & \textbf{11.0} & \textbf{13.8} & \textbf{10.9} \\
    \bottomrule
    \end{tabular}%
    \end{threeparttable}
    \captionof{table}{Results of molecule proportion in terms of ring number (P), atom stability (AS), molecule validity (V), novelty (N), and success rate (S). The number of molecules with above 11 rings in GEOM-DRUG is lower than 100.}
    \label{tab: Drug}
\end{table}
\textbf{Ring-Structure Molecule Generation.} 

\textit{QM9 Dataset.} All models were trained on identical datasets containing molecules with 0–3 rings. We evaluated their in-distribution (0–3 rings) and out-of-distribution (OOD; 4–8 rings) generation performance. Results for 10,000 generated molecules per ring-count distribution are shown in Table~\ref{tab: ring coverage}. Atom stability, molecule stability, validity, novelty, and success rate are averaged across four training distributions and five target distributions (full results in Appendix~\ref{appendix: ring results}). \textit{1)} Table~\ref{tab: ring coverage} demonstrates that uncontrollable baseline methods (e.g., EDM, GeoLDM, EquiFM, GeoBFN) exhibit limited success in OOD generation, achieving $\leq 7.0\%$ success rates for 4–8 ring molecules (Table~\ref{tab: ring coverage}), signifying the challenge of OOD generation under the structural shift. \textit{2)} Incorporating explicit ring-count controls (C-EDM, C-GeoLDM, EEGSDE) and dedicated OOD methods (e.g., MOOD, CGD) moderately improves OOD performance (up to 26.2\%), underscoring the difference between property shift and structural shift and the inherent difficulty of the latter. \textit{3)} Fragment-based approaches (DiffLinker, LinkerNet), while using identical input data as our method, underperform across all metrics, particularly in OOD settings. This highlights both the limitations of existing fragment-based frameworks in generalization and the critical role of our distributional prior representations in guiding molecular generative models. \textit{4)} Our~\textit{GODD} achieves a 40.5\% success rate. Moreover, we observe that most methods (i.e., C-GeoLDM, C-EDM, and EEGSDE) and OOD methods (MOOD and CGD) cannot generate 8-ring molecules, reflecting the difficulty of generating those complex and sparse molecules in the original QM9 (only 36 8-ring molecules). Our results validate that \textit{GODD} achieves robust OOD 3D molecule generation, even under challenging ring-structure shifts and data-sparse distributional regimes. These findings underscore the necessity of domain-specific inductive biases for generalizable molecular generation in low-data settings.


\textit{GEOM-DRUG Dataset.} Table~\ref{tab: Drug} presents statistical comparisons of molecular generation methods for rare ring systems (11–14 and 22 rings) on the GEOM-DRUG dataset, where molecules with higher ring counts exhibit extreme sparsity. Both unconditional baselines (EDM, GeoLDM, EquiFM, and GeoBFN) and OOD baselines (MOOD and CGD) fail to generate molecules exceeding 10 rings. Although DiffLinker and LinkerNet incorporate OOD structural priors, they underperform \textit{GODD} across all metrics, including proportion (p $<$ 0.1015), validity (11.0\% vs. 5.01\% and 7.92\%), and novelty (13.8\% vs. 6.25\% and 9.73\%). These results further confirm that fragment-based drug design methods inherently lack OOD generalization capabilities. In contrast, \textit{GODD} achieves an average success rate of 13.8\% for OOD ring counts despite being trained exclusively on molecules containing 0–10 rings, demonstrating robust generalization.

\begin{table*}[t]
\small
  \centering
  \begin{threeparttable}
  \begin{adjustbox}{width=1\textwidth,left}
    \begin{tabular}{c|ccccc|ccccc|ccccc}
    \toprule
    Distributions & \multicolumn{5}{c|}{In Distribution (\%)} & \multicolumn{5}{c|}{OOD I (\%)}        & \multicolumn{5}{c}{OOD II (\%)} \\
    \midrule
    Metric $\uparrow$ & P     & C     & V     & N     & S     & P     & C     & V     & N     & S     & P     & C     & V     & N     & S \\
    \midrule
    Data  & 76.4  & 100.0 & 97.7  & -     & -     & 11.5  & 100.0 & 97.7  & -     & -     & 12.1  & 100.0 & 97.7  & -     & - \\
    \midrule
    EDM & 91.4  & 56.5  & 83.2  & 58.2  & 52.0  & 5.9   & 26.5  & 5.3   & 3.7   & 3.3   & 2.7   & 17.0  & 2.4   & 1.7   & 1.5 \\
    GeoLDM & 90.6  & 54.3  & 81.7  & 57.8  & 51.0  & 5.9   & 26.7  & 5.3   & 3.8   & 3.3   & 3.5   & 19.0  & 3.2   & 2.3   & 2.0 \\
    EquiFM & 91.0  & 56.3  & 86.2  & 48.9  & 46.3  & 5.4   & 27.8  & 5.1   & 2.9   & 2.7   & 3.6   & 17.4  & 0.0   & 0.0   & 0.0 \\
    GeoBFN & 91.1  & 54.4  & 86.8  & 60.5  & 57.7  & 6.0   & 27.3  & 5.7   & 4.0   & 3.8   & 2.9   & 19.9  & 2.7   & 1.9   & 1.8 \\
    \midrule
    DiffLinker & 91.8  & 74.8  & 83.9  & 52.8  & 47.7  & 90.3  & 81.8  & 79.0  & 79.9  & 58.7  & 80.4  & 60.1  & 66.2  & 59.7  & 42.8 \\
    LinkerNet & 89.4  & 74.5  & 82.1  & 62.5  & 49.9  & 87.9  & 83.0  & 76.2  & 75.8  & 64.3  & 79.9  & 61.0  & 65.2  & 58.0  & 53.2 \\
    \midrule
    \rowcolor{lightgray}
    \textit{GODD} & \textbf{99.2} & \textbf{92.5}  & \textbf{90.7} & \textbf{67.6} & \textbf{52.4} & \textbf{97.0} & \textbf{97.1} & \textbf{80.0} & \textbf{84.5} & \textbf{68.9} & \textbf{95.5} & \textbf{85.7} & \textbf{83.3} & \textbf{82.0} & \textbf{65.8} \\
    \bottomrule
    \end{tabular}%
    \end{adjustbox}
    \end{threeparttable}
    \caption{Results of proportion (P), scaffold coverage (C), molecule validity (V), molecule novelty (N), and success rate (S).}
  \label{tab: scaffold}%
\end{table*}%

\begin{table}[t]
    \centering
    \begin{tabular}{c|cc|cc|cc}
        \toprule
        Distributions & \multicolumn{2}{c|}{In-dist (\%)} & \multicolumn{2}{c|}{OOD I (\%)} & \multicolumn{2}{c}{OOD II (\%)} \\
        \midrule
        \# Metric$\uparrow$ & AS    & MS    & AS    & MS    & AS    & MS \\
        \midrule
        Data  & 99.0  & 95.2  & 99.0  & 95.2  & 99.0  & 95.2 \\
        \midrule
        EDM   & 90.4  & 73.3  & 5.8   & 4.7   & 2.6   & 2.1 \\
        GeoLDM & 89.1  & 75.6  & 5.8   & 4.9   & 3.5   & 3.0 \\
        EquiFM & 90.0  & 80.4  & 5.3   & 4.8   & 3.6   & 3.2 \\
        GeoBFN & 90.3  & \textbf{82.8} & 5.9   & 5.5   & 2.9   & 2.6 \\
        \midrule
        DiffLinker & 89.6  & 57.5  & 85.6  & 29.5  & 76.5  & 30.3 \\
        LinkerNet & 87.3  & 69.2  & 83.5  & 37.3  & 71.5  & 30.6 \\
        \midrule
        \rowcolor{lightgray}
        \textit{GODD} & \textbf{96.1} & 71.3  & \textbf{89.5} & \textbf{45.6} & \textbf{89.0} & \textbf{35.1}\\
        \bottomrule
        \end{tabular}%
    \caption{Results of atom stability and molecule stability.}
    \label{tab: scaffold-as-ms}
\end{table}

\begin{figure}
    \centering
    \includegraphics[width=0.9\linewidth]{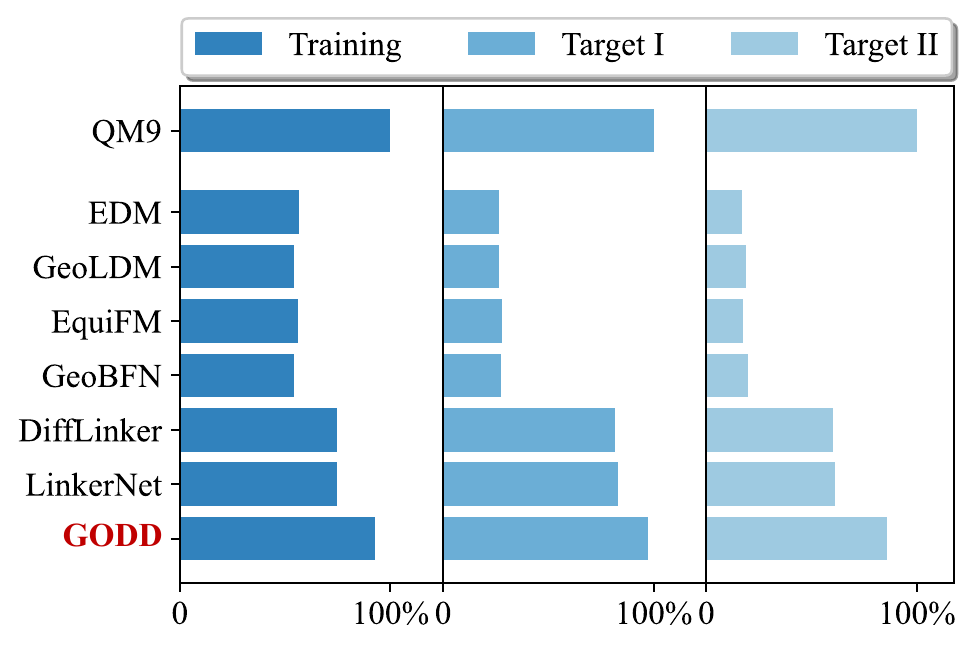}
    \caption{Visualization of Coverage. Molecules generated by the proposed method cover more OOD scaffolds.}
    \label{fig: qm9 scaffold proportion and coverage}
\end{figure}

\paragraph{Scaffold Molecule Generation.} In the task of OOD scaffold molecule generation, the structural sparsity of scaffolds (12,075 unique scaffolds among 15,831 molecules) precludes effective classifier training or the definition of specific scalar properties for conditional or existing OOD generative models. Consequently, conditional and OOD-specific frameworks are excluded from consideration. All baseline methods were trained exclusively on in-distribution data. Upon completion of training, each model generated 15,000 molecules under in-distribution, OOD-I, and OOD-II settings. Quantitative results across evaluation metrics are presented in Table~\ref{tab: scaffold}, Table~\ref{tab: scaffold-as-ms}, and Figure~\ref{fig: qm9 scaffold proportion and coverage}. Baseline methods (EDM, GeoLDM, EquiFM, GeoBFN) produce molecules with scaffold proportions closely resembling the training distribution but fail to adequately approximate the target OOD-I and OOD-II distributions (Table~\ref{tab: scaffold}). Fragment-based methods (DiffLinker, LinkerNet) achieve reasonable in-distribution performance but demonstrate marked degradation in OOD generation, with a 12\% reduction in scaffold coverage for the highly sparse OOD-II scaffolds. This further corroborates the limitations of fragment-based approaches in OOD scenarios. In contrast, our proposed \textit{GODD} framework successfully generates OOD molecules containing specified target scaffolds when provided with molecular substructures, achieving scaffold proportions exceeding 95\% for both OOD distributions. The method demonstrates consistent performance across in-distribution and OOD settings, validating the efficacy of leveraging distributional priors to steer 3D molecular generation.

Notably, for OOD-II—which comprises over 12,000 distinct rare scaffolds—only \textit{GODD} achieves 85.7\% scaffold coverage, underscoring the critical contribution of our \textit{EAAE}. \textit{GODD} operates without requiring OOD training data, instead leveraging molecular substructures as structural priors to circumvent data scarcity constraints. Under the same condition with baselines DiffLinker and LinkerNet, the framework achieves improvements of up to 22.3\% in molecular novelty and 12.6\% in success rate relative to fragment-based baselines. As evidenced by the atom- and molecule-level stability metrics in Table~\ref{tab: scaffold-as-ms}, \textit{GODD} demonstrates superior capability in generating chemically stable molecules with target scaffolds compared to existing approaches.


\begin{table}[t]
    \centering
    \begin{adjustbox}{width=1\linewidth,center}
    \begin{threeparttable}
    \begin{tabular}{c|cccc}
    \toprule
          GEOM-LINKER & QED $\uparrow$  & SA $\downarrow$ & V (\%) $\uparrow$ & S (\%) $\uparrow$ \\
    \midrule
    DiffLinker & 0.56  & 3.92 & 42.17 & 14.45 \\
    LinkerNet &  0.56 & 3.89 &  48.5 & 18.9 \\
    \midrule
    \rowcolor{lightgray}
    \textit{GODD}  & \textbf{0.57} & \textbf{3.63}  & \textbf{65.2} & \textbf{22.61} \\
    \bottomrule
    \end{tabular}%
    \end{threeparttable}
    \end{adjustbox}
    \captionof{table}{Results on the quantitative estimate of drug-likeness (QED), synthetic accessibility (SA), validity (v), and success rate (S) on the OOD linker design task.}
    \label{tab: linker-design}
\end{table}
\textbf{Evaluation on the Task of Linker Design.} 
Beyond validity and uniqueness, we incorporate established metrics from prior work, including quantitative drug-likeness (QED) and synthetic accessibility (SA) scores. Benchmark analyses reveal that existing linker design methods underperform in out-of-distribution (OOD) fragment linking, with validity rates below 50\%. In contrast, our framework achieves 65.2\% validity. Furthermore, our approach yields molecules with higher QED and lower SA scores, demonstrating advantages in both drug-likeness and synthesizability. These results demonstrate \textit{GODD} as a robust solution for OOD fragment linking, surpassing conventional methods in critical performance metrics.

\begin{table}[t]
    \centering
    \begin{adjustbox}{width=1\linewidth,center}
    \begin{threeparttable}
    \begin{tabular}{c|ccc|ccc|ccc}
    \toprule
    \multicolumn{1}{c|}{Distribution} & \multicolumn{3}{c|}{In-dist (\%)} & \multicolumn{3}{c|}{OOD I (\%)} & \multicolumn{3}{c}{OOD II (\%)} \\
    \midrule
    \multicolumn{1}{c|}{Metrics$\uparrow$} & P     & C     & V     & P     & C     & V     & P     & C     & V \\
    \midrule
    \textit{GODD}* & \textbf{99.2} & \textbf{98.5} & 85.1  & 95.1  & 96.9  & 58.3  & 94.3  & 84.0  & 35.0 \\
    \textit{GODD} & \textbf{99.2} & 92.5  & \textbf{90.7} & \textbf{97.0} & \textbf{97.1} & \textbf{80.0} & \textbf{95.5} & \textbf{85.7} & \textbf{83.3} \\
    \midrule
    \multicolumn{1}{c|}{Metrics$\uparrow$} & AS    & MS    & S     & AS    & MS    & S     & AS    & MS    & S \\
    \midrule
    \textit{GODD}* & 89.2  & 68.4  & 52.1  & 82.0  & 12.8  & 41.8  & 75.1  & 10.4  & 31.0 \\
    \textit{GODD} & \textbf{96.1} & \textbf{71.3}  & \textbf{52.4} & \textbf{89.5} & \textbf{45.6} & \textbf{68.9} & \textbf{89.0} & \textbf{35.1} & \textbf{65.8} \\
    \bottomrule
    \end{tabular}
    \begin{tablenotes}
    \small
        \item[$*$] \textit{GODD}* with a \textit{symmetric autoencoder}.
    \end{tablenotes}
    \end{threeparttable}
    \end{adjustbox}
    \captionof{table}{Results of proportion (P), scaffold coverage (C), molecule validity (V), molecule success rate (S), atom stability (AS), and molecule stability (MS).}
    \label{tab: ablation}
\end{table}
\textbf{Ablation Study for Evaluating the Significance of the Asymmetric Autoencoder.} We present the ablation study in Table~\ref{tab: ablation} featuring a variation of the proposed method, \textit{GODD}*, which utilizes a \textit{symmetric autoencoder}. Specifically, the autoencoder of \textit{GODD} receives and reconstructs only the substructure. The results indicate that \textit{GODD}* demonstrates promising in-distribution generation and achieves better performance in scaffold coverage, aligning with the performance of traditional autoencoders in the in-distribution tasks. However, \textit{GODD}* performs worse than \textit{GODD} in OOD generation. Although \textit{GODD}* achieves similar proportions and coverage by receiving OOD substructures, its generation quality is worse, particularly regarding stability and validity. This suggests that even with substructures, \textit{GODD}* is still hard to generalize to generate valid molecules in OOD scenarios. These observations underscore the effectiveness of using asymmetric autoencoder.
\section{Conclusion}
This paper investigated the problem of OOD molecule generation in the context of structural shifts and proposed an asymmetric autoencoder to represent substructures as structural priors to steer the generation toward data-sparse regions.
Our quantitative experiments demonstrated that the proposed method surpasses existing techniques, including unconditional, conditional, and OOD approaches, in generating valid, unique, and novel OOD molecules with desired substructures in data-sparse regions. Extensive quantitative results in successful OOD generation validated the ability of asymmetric autoencoder to encode unseen structure and the potential of \textit{GODD} in steering generation through the encoded structural priors. Furthermore, the linker design experiment confirmed the proposed method's applicability to fragment-based drug design. Additionally, our framework is generative model-agnostic; it can be seamlessly integrated into other generative models, such as latent diffusion~\cite{pmlr-v202-xu23n} or flow-based models~\cite{equifm}.
\clearpage
\section*{Acknowledgments}
This work was partially supported by the Research Grants Council (RGC) of the Hong Kong (HK) SAR (Grant No. 15208725 and 15208222), the Young Scientists Fund of National Natural Science Foundation of China (NSFC) (Grant No. 62206235), and the Hong Kong Polytechnic University (Grant No. A0046682 and P0057774).
\bibliography{damg}

@InProceedings{pmlr-v37-sohl-dickstein15,
  title =    {{Deep Unsupervised Learning using Nonequilibrium Thermodynamics}},
  author =   {Sohl-Dickstein, Jascha and Weiss, Eric and Maheswaranathan, Niru and Ganguli, Surya},
  booktitle =    {Proceedings of the 32nd International Conference on Machine Learning},
  pages =    {2256--2265},
  year =     {2015},
  editor =   {Bach, Francis and Blei, David},
  volume =   {37},
  publisher =    {PMLR}
}

@inproceedings{NEURIPS2020_4c5bcfec,
 author = {Ho, Jonathan and Jain, Ajay and Abbeel, Pieter},
 booktitle = {Advances in Neural Information Processing Systems},
 pages = {6840--6851},
 title = {{Denoising Diffusion Probabilistic Models}},
 volume = {33},
 year = {2020}
}

@article{doi:10.1021/jm9602928,
author = {Bemis, Guy W. and Murcko, Mark A.},
title = {{The Properties of Known Drugs. 1. Molecular Frameworks}},
journal = {Journal of Medicinal Chemistry},
volume = {39},
number = {15},
pages = {2887-2893},
year = {1996}

}

@book{serre1977linear,
  title={{Linear Representations of Finite Groups}},
  author={Serre, Jean-Pierre and others},
  volume={42},
  year={1977},
  publisher={Springer}
}

@InProceedings{pmlr-v162-hoogeboom22a,
  title =    {{Equivariant Diffusion for Molecule Generation in 3D}},
  author =       {Hoogeboom, Emiel and Satorras, V\'{\i}ctor Garcia and Vignac, Cl{\'e}ment and Welling, Max},
  booktitle =    {Proceedings of the 39th International Conference on Machine Learning},
  pages =    {8867--8887},
  year =     {2022},
  volume =   {162},
  month =    {17--23 Jul},
  publisher =    {PMLR}
}

@article{RN363,
   author = {Watson, Joseph L. and Juergens, David and Bennett, Nathaniel R. and Trippe, Brian L. and Yim, Jason and Eisenach, Helen E. and Ahern, Woody and Borst, Andrew J. and Ragotte, Robert J. and Milles, Lukas F. and Wicky, Basile I. M. and Hanikel, Nikita and Pellock, Samuel J. and Courbet, Alexis and Sheffler, William and Wang, Jue and Venkatesh, Preetham and Sappington, Isaac and Torres, Susana Vázquez and Lauko, Anna and De Bortoli, Valentin and Mathieu, Emile and Ovchinnikov, Sergey and Barzilay, Regina and Jaakkola, Tommi S. and DiMaio, Frank and Baek, Minkyung and Baker, David},
   title = {{De Novo Design of Protein Structure and Function with RFdiffusion}},
   journal = {Nature},
   volume = {620},
   number = {7976},
   pages = {1089-1100},
   ISSN = {1476-4687},
   year = {2023}
}

@inproceedings{xie2022crystal,
title={{Crystal Diffusion Variational Autoencoder for Periodic Material Generation}},
author={Tian Xie and Xiang Fu and Octavian-Eugen Ganea and Regina Barzilay and Tommi S. Jaakkola},
booktitle={International Conference on Learning Representations},
year={2022}
}

@inproceedings{luo2022an,
title={{An Autoregressive Flow Model for 3D Molecular Geometry Generation from Scratch}},
author={Youzhi Luo and Shuiwang Ji},
booktitle={International Conference on Learning Representations},
year={2022}
}

@inproceedings{NEURIPS2021_21b5680d,
 author = {Garcia Satorras, Victor and Hoogeboom, Emiel and Fuchs, Fabian and Posner, Ingmar and Welling, Max},
 booktitle = {Advances in Neural Information Processing Systems},
 pages = {4181--4192},
 publisher = {Curran Associates, Inc.},
 title = {{E(n) Equivariant Normalizing Flows}},
 volume = {34},
 year = {2021}
}

@article{celeghini1991three,
  title={{The Three-Dimensional Euclidean Quantum Group E(3) Q and Its R-Matrix}},
  author={Celeghini, Enrico and Giachetti, Riccardo and Sorace, Emanuele and Tarlini, Marco},
  journal={Journal of Mathematical Physics},
  volume={32},
  number={5},
  pages={1159--1165},
  year={1991},
  publisher={American Institute of Physics}
}

@InProceedings{exploring-ood,
  title =    {{Exploring Chemical Space with Score-Based Out-of-distribution Generation}},
  author =       {Lee, Seul and Jo, Jaehyeong and Hwang, Sung Ju},
  booktitle =    {Proceedings of the 40th International Conference on Machine Learning},
  pages =    {18872--18892},
  year =     {2023},
  volume =   {202},
  month =    {23--29 Jul},
  publisher =    {PMLR}
}

@InProceedings{context-ood,
  title={Context-Guided Diffusion for Out-of-Distribution Molecular and Protein Design},
  author={Klarner, Leo and Rudner, Tim GJ and Morris, Garrett M and Deane, Charlotte M and Teh, Yee Whye},
  booktitle = {Proceedings of the 41th International Conference on Machine Learning},
  year={2024}
}

@inproceedings{NEURIPS2021_41da609c,
 author = {Yang, Soojung and Hwang, Doyeong and Lee, Seul and Ryu, Seongok and Hwang, Sung Ju},
 booktitle = {Advances in Neural Information Processing Systems},
 pages = {7924--7936},
 publisher = {Curran Associates, Inc.},
 title = {{Hit and Lead Discovery with Explorative RL and Fragment-based Molecule Generation}},
 volume = {34},
 year = {2021}
}

@article{RN364,
   author = {Walters, W. Patrick and Murcko, Mark},
   title = {{Assessing the Impact of Generative AI on Medicinal Chemistry}},
   journal = {Nature Biotechnology},
   volume = {38},
   number = {2},
   pages = {143-145},
   ISSN = {1546-1696},
   year = {2020},
   type = {Journal Article}
}

@InProceedings{He_2022_CVPR,
    author    = {He, Kaiming and Chen, Xinlei and Xie, Saining and Li, Yanghao and Doll\'ar, Piotr and Girshick, Ross},
    title     = {{Masked Autoencoders Are Scalable Vision Learners}},
    booktitle = {Proceedings of the IEEE/CVF Conference on Computer Vision and Pattern Recognition (CVPR)},
    month     = {June},
    year      = {2022},
    pages     = {16000-16009}
}

@InProceedings{pmlr-v80-jin18a,
  title =    {{Junction Tree Variational Autoencoder for Molecular Graph Generation}},
  author =       {Jin, Wengong and Barzilay, Regina and Jaakkola, Tommi},
  booktitle =    {Proceedings of the 35th International Conference on Machine Learning},
  pages =    {2323--2332},
  year =     {2018},
  publisher =    {PMLR}

}

@inproceedings{Shi*2020GraphAF:,
    title={{GraphAF: a Flow-Based Autoregressive Model for Molecular Graph Generation}},
    author={Chence Shi and Minkai Xu and Zhaocheng Zhu and Weinan Zhang and Ming Zhang and Jian Tang},
    booktitle={International Conference on Learning Representations},
    year={2020}
}

@inproceedings{NEURIPS2018_b8a03c5c,
 author = {Liu, Qi and Allamanis, Miltiadis and Brockschmidt, Marc and Gaunt, Alexander},
 booktitle = {Advances in Neural Information Processing Systems},
 editor = {S. Bengio and H. Wallach and H. Larochelle and K. Grauman and N. Cesa-Bianchi and R. Garnett},
 pages = {},
 publisher = {Curran Associates, Inc.},
 title = {{Constrained Graph Variational Autoencoders for Molecule Design}},
 volume = {31},
 year = {2018}
}

@inproceedings{NEURIPS2019_a4d8e2a7,
 author = {Gebauer, Niklas and Gastegger, Michael and Sch\"{u}tt, Kristof},
 booktitle = {Advances in Neural Information Processing Systems},
 publisher = {Curran Associates, Inc.},
 title = {{Symmetry-Adapted Generation of 3D Point Sets for The Targeted Discovery of Molecules}},
 volume = {32},
 year = {2019}
}

@InProceedings{pmlr-v202-xu23n,
  title={{Geometric Latent Diffusion Models for 3D Molecule Generation}},
  author={Xu, Minkai and Powers, Alexander S and Dror, Ron O and Ermon, Stefano and Leskovec, Jure},
  booktitle={International Conference on Machine Learning},
  pages={38592--38610},
  year={2023},
  organization={PMLR}
}

@inproceedings{
wu2022diffusionbased,
title={{Diffusion-Based Molecule Generation with Informative Prior Bridges}},
author={Lemeng Wu and Chengyue Gong and Xingchao Liu and Mao Ye and qiang liu},
booktitle={Advances in Neural Information Processing Systems},
editor={Alice H. Oh and Alekh Agarwal and Danielle Belgrave and Kyunghyun Cho},
year={2022},
}

@inproceedings{
Song2024unified,
title={{Unified Generative Modeling of 3D Molecules with Bayesian Flow Networks}},
author={Yuxuan Song and  J. Gong and Y. Qu and M. Zheng and H. Zhou and J. Liu and Wei-Ying Ma},
booktitle={The Twelfth International Conference on Learning Representations},
year={2024}
}

@article{peng2023moldiff,
  title={{MolDiff: Addressing the Atom-Bond Inconsistency Problem in 3D Molecule Diffusion Generation}},
  author={Peng, Xingang and Guan, Jiaqi and Liu, Qiang and Ma, Jianzhu},
  journal={arXiv preprint arXiv:2305.07508},
  year={2023}
}

@article{MoleculeNet,
   author = {Wu, Z. and Ramsundar, B. and Feinberg, E. N. and Gomes, J. and Geniesse, C. and Pappu, A. S. and Leswing, K. and Pande, V.},
   title = {{MoleculeNet: A Benchmark for Molecular Machine Learning}},
   journal = {Chem Sci},
   volume = {9},
   number = {2},
   pages = {513-530},
   ISSN = {2041-6520 (Print)
2041-6520},
   DOI = {10.1039/c7sc02664a},
   year = {2018},
   type = {Journal Article}
}

@article{Simon_ring_number,
author = {Simon E Ward and Paul Beswick},
title = {{What Does the Aromatic Ring Number Mean for Drug Design?}},
journal = {Expert Opinion on Drug Discovery},
volume = {9},
number = {9},
pages = {995-1003},
year = {2014},
publisher = {Taylor & Francis},
doi = {10.1517/17460441.2014.932346},
    note ={PMID: 24955724},
}

@article{RITCHIE20091011,
title = {{The Impact of Aromatic Ring Count on Compound Developability – Are Too Many Aromatic Rings A Liability in Drug Design?}},
journal = {Drug Discovery Today},
volume = {14},
number = {21},
pages = {1011-1020},
year = {2009},
issn = {1359-6446},
author = {Timothy J. Ritchie and Simon J.F. Macdonald}
}

@article{doi:10.1021/ci300415d,
author = {Ruddigkeit, Lars and van Deursen, Ruud and Blum, Lorenz C. and Reymond, Jean-Louis},
title = {{Enumeration of 166 Billion Organic Small Molecules in the Chemical Universe Database GDB-17}},
journal = {Journal of Chemical Information and Modeling},
volume = {52},
number = {11},
pages = {2864-2875},
year = {2012},
doi = {10.1021/ci300415d},
    note ={PMID: 23088335},
}

@InProceedings{pmlr-v139-satorras21a,
  title = 	 {{E(n) Equivariant Graph Neural Networks}},
  author =       {Satorras, V\'{\i}ctor Garcia and Hoogeboom, Emiel and Welling, Max},
  booktitle = 	 {Proceedings of the 38th International Conference on Machine Learning},
  pages = 	 {9323--9332},
  year = 	 {2021},
  editor = 	 {Meila, Marina and Zhang, Tong},
  volume = 	 {139},
  month = 	 {18--24 Jul},
  publisher =    {PMLR}
}

@article{landrum2016rdkit,
  title={{Rdkit: Open-Source Cheminformatics Software}},
  author={Landrum, Greg and others}, 
  journal={Open-source Cheminformatics},
  year={2016}
}

@inproceedings{kingma2013auto,
  title={{Auto-Encoding Variational Bayes}},
  author={Kingma, Diederik P. and Welling, Max},
  booktitle={2nd International Conference on Learning Representations},
  year={2013}
}

@article{QM9,
   author = {Ramakrishnan, Raghunathan and Dral, Pavlo O. and Rupp, Matthias and von Lilienfeld, O. Anatole},
   title = {{Quantum Chemistry Structures and Properties of 134 Kilo Molecules}},
   journal = {Scientific Data},
   volume = {1},
   number = {1},
   pages = {140022},
   ISSN = {2052-4463},
   year = {2014},
   type = {Journal Article}
}

@inproceedings{DBLP:journals/corr/KingmaW13,
  author       = {Diederik P. Kingma and
                  Max Welling},
  title        = {{Auto-Encoding Variational Bayes}},
  booktitle    = {2nd International Conference on Learning Representations, {ICLR} 2014,
                  Banff, AB, Canada, April 14-16, 2014, Conference Track Proceedings},
  year         = {2014},
}

@article{RN367,
   author = {Karageorgis, George and Warriner, Stuart and Nelson, Adam},
   title = {{Efficient Discovery of Bioactive Scaffolds by Activity-Directed Synthesis}},
   journal = {Nature Chemistry},
   volume = {6},
   number = {10},
   pages = {872-876},
   ISSN = {1755-4349},
   DOI = {10.1038/nchem.2034},
   year = {2014},
   type = {Journal Article}
}

@article{geomdrug,
   author = {Axelrod, Simon and Gómez-Bombarelli, Rafael},
   title = {{GEOM, Energy-Annotated Molecular Conformations for Property Prediction and Molecular Generation}},
   journal = {Scientific Data},
   volume = {9},
   number = {1},
   pages = {185},
   ISSN = {2052-4463},   year = {2022},
   type = {Journal Article}
}

@inproceedings{
bao2023equivariant,
title={{Equivariant Energy-Guided {SDE} for Inverse Molecular Design}},
author={Fan Bao and Min Zhao and Zhongkai Hao and Peiyao Li and Chongxuan Li and Jun Zhu},
booktitle={The Eleventh International Conference on Learning Representations},
year={2023},
}

@InProceedings{KingBa15,
  author    = {Kingma, Diederik and Ba, Jimmy},
  booktitle = {International Conference on Learning Representations (ICLR)},
  title     = {{Adam: A Method for Stochastic Optimization}},
  year      = {2015},
  address   = {San Diega, CA, USA},
  optmonth  = {12},
}

@article{equifm,
  title={{Equivariant Flow Matching with Hybrid Probability Transport for 3D Molecule Generation}},
  author={Song, Yuxuan and Gong, Jingjing and Xu, Minkai and Cao, Ziyao and Lan, Yanyan and Ermon, Stefano and Zhou, Hao and Ma, Wei-Ying},
  journal={Advances in Neural Information Processing Systems},
  volume={36},
  year={2023}
}

@inproceedings{geobfn,
  title={{Unified Generative Modeling of 3D Molecules with Bayesian Flow Networks}},
  author={Song, Yuxuan and Gong, Jingjing and Zhou, Hao and Zheng, Mingyue and Liu, Jingjing and Ma, Wei-Ying},
  booktitle={The Twelfth International Conference on Learning Representations},
  year={2023}
}

@article{linker-design-NMI,
  title={{Equivariant 3D-Conditional Diffusion Model for Molecular Linker Design}},
  author={Igashov, Ilia and St{\"a}rk, Hannes and Vignac, Cl{\'e}ment and Schneuing, Arne and Satorras, Victor Garcia and Frossard, Pascal and Welling, Max and Bronstein, Michael and Correia, Bruno},
  journal={Nature Machine Intelligence},
  pages={1--11},
  year={2024},
  publisher={Nature Publishing Group UK London}
}

@article{DA-mol-learning,
  title={{Learning Invariant Molecular Representation in Latent Discrete Space}},
  author={Zhuang, Xiang and Zhang, Qiang and Ding, Keyan and Bian, Yatao and Wang, Xiao and Lv, Jingsong and Chen, Hongyang and Chen, Huajun},
  journal={Advances in Neural Information Processing Systems},
  volume={36},
  pages={78435--78452},
  year={2023}
}

@article{linkernet-nips,
  title={{LinkerNet: Fragment Poses and Linker Co-design with 3D Equivariant Diffusion}},
  author={Guan, Jiaqi and Peng, Xingang and Jiang, Peiqi and Luo, Yunan and Peng, Jian and Ma, Jianzhu},
  journal={Advances in Neural Information Processing Systems},
  volume={36},
  year={2024}
}

@article{fbdd-rise,
  title={{The rise of fragment-based drug discovery}},
  author={Murray, Christopher W and Rees, David C},
  journal={Nature Chemistry},
  volume={1},
  number={3},
  pages={187--192},
  year={2009},
  publisher={Nature Publishing Group}
}

@article{rule-3-for-fragment,
  title={{A'rule of three'for fragment-based lead discovery?}},
  author={Congreve, Miles and Carr, Robin and Murray, Chris and Jhoti, Harren},
  journal={Drug Discovery Today},
  volume={8},
  number={19},
  pages={876--877},
  year={2003}
}

@article{computational-fbdd,
  title={{Computational fragment-based drug design: current trends, strategies, and applications}},
  author={Bian, Yuemin and Xie, Xiang-Qun},
  journal={The AAPS Journal},
  volume={20},
  pages={1--11},
  year={2018},
  publisher={Springer}
}

@article{smiles-fbdd,
  title={{t-SMILES: A Fragment-based Molecular Representation Framework for de novo Ligand Design}},
  author={Wu, Juan-Ni and Wang, Tong and Chen, Yue and Tang, Li-Juan and Wu, Hai-Long and Yu, Ru-Qin},
  journal={Nature Communications},
  volume={15},
  number={1},
  pages={4993},
  year={2024},
  publisher={Nature Publishing Group UK London}
}

@article{unified-guidance-fbdd,
  title={{Unified Guidance for Geometry-Conditioned Molecular Generation}},
  author={Ayadi, Sirine and Hetzel, Leon and Sommer, Johanna and Theis, Fabian and G{\"u}nnemann, Stephan},
  journal={Advances in Neural Information Processing Systems},
  volume={37},
  pages={138891--138924},
  year={2024}
}

@InProceedings{diffpc,
    author    = {Luo, Shitong and Hu, Wei},
    title     = {{Diffusion Probabilistic Models for 3D Point Cloud Generation}},
    booktitle = {Proceedings of the IEEE/CVF Conference on Computer Vision and Pattern Recognition (CVPR)},
    month     = {June},
    year      = {2021},
    pages     = {2837-2845}
}

@inproceedings{
    hong2025accelerating,
    title={{Accelerating 3D Molecule Generation via Jointly Geometric Optimal Transport}},
    author={Haokai Hong and Wanyu Lin and Kay Chen Tan},
    booktitle={The Thirteenth International Conference on Learning Representations},
    year={2025}
}


\appendix
\clearpage
\onecolumn
\section*{Technical Appendices and Supplementary Material}

\newcommand{\supplementtoc}{%
    \renewcommand{\contentsname}{Supplementary Material Contents}%
    \startcontents[supp]
    \printcontents[supp]{}{1}{}
}

\supplementtoc


\section{Diffusion Models}
\label{appendix: dm detail}

{\bf Diffusion Models.} Diffusion models~\cite{pmlr-v37-sohl-dickstein15, diffpc} are latent variable models for learning distributions by modeling the reverse of a diffusion process~\cite{NEURIPS2020_4c5bcfec}. Given a data point $\mathbf{x}_{0}\sim q(\mathbf{x}_0)$ and a variance schedule $\beta_1,\dots,\beta_T$ that controls the amount of noise added at each timestep $t$, the diffusion process or forward process gradually add Gaussian noise to the data point $\mathbf{x}$:
\begin{equation}
    q(\mathbf{x}_{t}|\mathbf{x}_{t-1}) := \mathcal{N}(\mathbf{x}_{t}; \sqrt{1-\beta_{t}}\mathbf{x}_{t-1}, \beta_{t}\mathbf{I}).
\end{equation}
Generally, the diffusion process $q$ has no trainable parameters. The denoising process or reverse process aims at learning a parameterized generative process, which incrementally denoise the noisy variables $\mathbf{x}_{T:1}$ to approximately restore the data point $\mathbf{x}_{0}$ in the original data distribution: 
\begin{equation}
    p_{\theta}(\mathbf{x}_{t-1}|\mathbf{x}_{t}):=\mathcal{N}(\mathbf{x}_{t-1};\mathbf{\mu_{\theta}}(\mathbf{x}_{t},t),\mathbf{\Sigma_{\theta}}(\mathbf{x}_{t},t)),
\end{equation}
where the initial distribution $p(\mathbf{x}_{t})$ is sampled from standard Gaussian noise $\mathcal{N}(0,\mathbf{I})$.
The loss for training diffusion model $\mathcal{L}_{\textrm{DM}}:=\mathcal{L}_{t}$ is simplified as:
$\mathcal{L}_{\textrm{DM}}=\mathbb{E}_{\mathbf{x}_{0},\epsilon,t}\left[\Vert\epsilon-\epsilon_{\theta}(\mathbf{x}_{t},t)\Vert^{2} \right]$, where $w(t)=\frac{\beta_{t}}{2\sigma_{t}^{2}\alpha_{t}(1-\bar\alpha_{t})}$ is the reweighting term and could be set as 1 with promising sampling quality, and $\mathbf{x}_{t}=\sqrt{\bar\alpha_{t}}\mathbf{x}_{0}+\sqrt{1-\bar\alpha_{t}}\epsilon$.

Given a data point $\mathbf{x}_{0}\sim q(\mathbf{x}_0)$ and a variance schedule $\beta_1,\dots,\beta_T$ that controls the amount of noise added at each timestep $t$, the diffusion process or forward process gradually add Gaussian noise to the data point $\mathbf{x}$:
\begin{equation}
    q(\mathbf{x}_{t}|\mathbf{x}_{t-1}) := \mathcal{N}(\mathbf{x}_{t}; \sqrt{1-\beta_{t}}\mathbf{x}_{t-1}, \beta_{t}\mathbf{I}),
\end{equation}
where $\beta_{1:T}$ are chosen such that data point $\mathbf{x}$ will approximately converge to standard Gaussian, \emph{i.e.}, $q(\mathbf{x}_{T})\approx\mathcal{N}(0,\mathbf{I})$. 
Generally, the diffusion process $q$ has no trainable parameters. The denoising process or reverse process aims at learning a parameterized generative process, which incrementally denoise the noisy variables $\mathbf{x}_{T:1}$ to approximate restore the data point $\mathbf{x}_{0}$ in the original data distribution:
\begin{equation}
    p_{\theta}(\mathbf{x}_{t-1}|\mathbf{x}_{t}):=\mathcal{N}(\mathbf{x}_{t-1};\mathbf{\mu_{\theta}}(\mathbf{x}_{t},t),\mathbf{\Sigma_{\theta}}(\mathbf{x}_{t},t)),
\end{equation}
where the initial distribution $p(\mathbf{x}_{t})$ is sampled from standard Gaussian noise $\mathcal{N}(0,\mathbf{I})$.
The means $\mu_{\theta}$ typically are neural networks such as U-Nets for images or Transformers for text.
\par
The forward process is $q(\mathbf{x}_{1:T}|\mathbf{x}_{0})$ is an approximate posterior to the Markov chain, and the reverse process $p_{\theta}(\mathbf{x}_{0:T})$ is optimized by a variational lower bound on the negative log-likelihood of $\mathbf{x}_{0}$ by:
\begin{align}
\label{eq: lvlb}
    & \mathbb{E}[-\log p_{\theta}(\mathbf{x}_{0})] \leq \mathbb{E}_{q}\left [ -\log \frac{p_{\theta}(\mathbf{x}_{0:T})}{q(\mathbf{x}_{1:T}|\mathbf{x}_{0})} \right] \\
    = & \mathbb{E}_{q}\left [ -\log p(\mathbf{x}_{T}) -\sum_{t\geq1}^{T}\log \frac{p_{\theta}(\mathbf{x}_{t-1}|\mathbf{x}_{t})}{q(\mathbf{x}_{t}|\mathbf{x}_{t-1})} \right],
\end{align}
which is $\mathcal{L}_{\textrm{vlb}}$. To efficiently train the diffusion models, further improvements come to term $\mathcal{L}_{\textrm{vlb}}$ by variance reduction, and thereby Eq. (\ref{eq: lvlb}) is rewritten as:
\begin{align}
\label{eq: dmloss}
    \mathcal{L}_{\textrm{vlb}}=\mathbb{E}_{q} [\mathcal{L}_{T} + \sum_{t=2}^{T}\mathcal{L}_{t} + \mathcal{L}_{0}]
\end{align}
where $\mathcal{L}_{T}=\log\frac{q(\mathbf{x}_{T}|\mathbf{x}_{0})}{p_{\theta}(\mathbf{x}_{T})}$, which models the distance between a standard normal distribution and the final latent variable $q(\mathbf{x}_{T} |\mathbf{x}_{0})$, since the approximate posterior $q$ has no learnable parameters, so $\mathcal{L}_{T}$ is a constant during training and can be ignored. $\mathcal{L}_{0} = -\log p_{\theta}(\mathbf{x}_{0}|\mathbf{x}_{1})$ models the likelihood of the data given $\mathbf{x}_{0}$, which is close to zero and ignored as well if $\beta_0\approx0$ and $\mathbf{x}_{0}$ is discrete.
\par
$\mathcal{L}_{t}$ in Eq. (\ref{eq: dmloss}) is the loss for the reverse process and is given by:
\begin{equation}
    \mathcal{L}_{t} = \sum_{t\geq2}^{T}\log \frac{q(\mathbf{x}_{t-1}|\mathbf{x}_{0},\mathbf{x}_{t})}{p_{\theta}(\mathbf{x}_{t-1}|\mathbf{x}_{t})}.
\end{equation}
While in this formulation the neural network directly predicts $\hat{\mathbf{x}_{0}}$, \cite{NEURIPS2020_4c5bcfec} found that optimization is easier when predicting the Gaussian noise instead. Intuitively, the network is trying to predict which part of the observation $\mathbf{x}_t$ is noise originating from the diffusion process, and which part corresponds to the underlying data point $\mathbf{x}_{0}$. Then sampling $\mathbf{x}_{t-1} \sim p_{\theta}(\mathbf{x}_{t-1}|\mathbf{x}_{t})$ is to compute
\begin{equation}
    \mathbf{x}_{t-1} = \frac{1}{\sqrt{\alpha_t}} \left (\mathbf{x}_{t} - \frac{\sqrt{\beta_{t}}}{\sqrt{1-\bar\alpha_t}}\mathbf{\epsilon}_{\theta}(\mathbf{x}_{t}, t) \right) + \sigma_{t}\mathbf{z},
\end{equation}
where $\alpha_t:=1-\beta_{t}$, $\bar\alpha_{t}:=\prod_{s=1}^{t}\alpha_{s}$, and $\mathbf{z}\sim\mathcal{N}(\mathbf{0},\mathbf{I})$. And thereby $\mathcal{L}_{\textrm{DM}}:=\mathcal{L}_{t}$ is simplified to:
\begin{equation}
    \label{eq: dm loss}
    \mathcal{L}_{\textrm{DM}}=\mathbb{E}_{\mathbf{x}_{0},\epsilon,t}\left[ 
    w(t)
    \Vert
    \epsilon-\epsilon_{\theta}(
    \mathbf{x}_{t}
    ,t)
    \Vert^{2}
    \right]
\end{equation}
where $w(t)=\frac{\beta_{t}}{2\sigma_{t}^{2}\alpha_{t}(1-\bar\alpha_{t})}$ is the reweighting term and could be simply set as 1 with promising sampling quality, and $\mathbf{x}_{t}=\sqrt{\bar\alpha_{t}}\mathbf{x}_{0}+\sqrt{1-\bar\alpha_{t}}\epsilon$.

\section{Model Architecture Details}
\label{appendix: emae}
\subsection{Equivaraint Masked Autoencoder}
In this work, \textit{EAAE} considers visible molecular structural geometries as point clouds, without specifying the connecting bonds. Therefore, in practice, we take the point clouds as fully connected graph $\mathcal{G}$ and model the interactions between all atoms $v_i \in \mathcal{V}$. Each node $v_i$ is embedded with coordinates $\mathbf{x}_{i} \in \mathbb{R}^{3}$ and atomic features $\mathbf{h}_{i}\in\mathbf{R}^{d}$. Then, \textit{EAAE} are composed of multiple Equivariant Convolutional Layers, and each single layer is expressed as~\cite{pmlr-v139-satorras21a}:
\begin{equation}
\label{eq: egnn}
    \begin{aligned}
        d_{ij}^{2}&=\Vert\mathbf{x}_{i}^{l}-\mathbf{x}_{j}^{l}\Vert^{2},\\
        \mathbf{m}_{i,j}&=\phi_{e}(\mathbf{h}_{i}^{l},\mathbf{h}_{j}^{l},d_{ij}^{2},a_{ij}),\\
        \mathbf{x}_{i}^{l+1}&=\mathbf{x}_{i}^{l}+\sum_{j\neq i}\frac{\mathbf{x}_{i}^{l}-\mathbf{x}_{j}^{l}}{d_{ij}+1}\phi_{x}(\mathbf{m}_{i,j})\\
        \mathbf{h}_{i}^{l+1}&=\phi_{h}(\mathbf{h}_{i}^{l},\sum_{j\in\mathcal{N}(i)}\phi_{i}(\mathbf{m}_{ij})\mathbf{m}_{ij})
    \end{aligned}
\end{equation}
where $l$ denotes the layer index, $\phi_{i}(\mathbf{m}_{ij})$ reweights messages passed from different edges in an attention weights manner, $d_{ij}+1$ is normalizing the relative directions $\mathbf{x}_{i}^{l}-\mathbf{x}_{j}^{l}$ following previous methods~\cite{pmlr-v139-satorras21a,pmlr-v162-hoogeboom22a}. All learnable functions, \textit{i.e.}, $\phi_{e}, \phi_{x}, \phi_{h}$, and, $\phi_{i}$, are parameterized by Multi Layer Perceptrons (MLPs). Then a complete EGNN model can be realized by stacking $L$ layers such that and satisfies the required equivariant constraint in Equations \ref{eq: encoder-equivariant}, \ref{eq: decoder-equivariant}, and \ref{eq: diffusion equivariant}.
\subsection{Equivaraint Structural Prior Steered Denoising Neural Networks}
\label{appendix: dsdm detail}
The denoising neural network is implemented by multiple equivariant convolutional layers, and the difference in the Equation~\ref{eq: egnn} is the hidden features $\mathbf{h}$. Due to the diffusion model is conditioned by $\mathbf{f}_{\mathbf{x}}, \mathbf{f}_{\mathbf{h}}$ from encoder $\mathcal{E}$, the hidden features for our denoising neural network is expressed as $\bar{\mathbf{h}} \gets [\mathbf{h}, \mathbf{f}_{\mathbf{x}}, \mathbf{f}_{\mathbf{h}}]$, where $\mathbf{h}$ are original features of geometric graph and $[a, b]$ is concatenation operation.
\subsection{Multi-Modal Feature Representation of Molecules}
\label{appendix: MMF}
Multimodal features of molecules raise concerns for the term $\mathcal{L}_{0}=-\log p_{\theta}(\mathbf{x}_{0}|\mathbf{x}_{1})$ in Equation \ref{eq: dmloss}. For categorical features such as the atom types, this model would however introduce an undesired bias~\cite{pmlr-v162-hoogeboom22a}. 
For the intermediate variable $\mathbf{x}_{t}$, we subdivide it into $\mathbf{z}_{\mathbf{x},t}$ and $\mathbf{z}_{\mathbf{h},t}$ in the proposed DM, which are coordinate variables and atomic attribute variables, respectively. 
\par
\textbf{Coordinate Features.} First we set $\sigma_{t}^{2}\mathbf{I}\gets \Sigma_{\theta}(\mathbf{x}_{t},t)=\beta_{t}$ and add an additional correction term containing the estimated noise $\boldsymbol{\epsilon}_{\mathbf{x},0}$ from denoising neural network $\boldsymbol{\epsilon}$. Then continuous positions $\mathbf{z}_{\mathbf{x}}$ in $p(\mathbf{z}_{\mathbf{x},0}|\mathbf{z}_{\mathbf{x},1})$ is expressed as:
\begin{equation}
    p(\mathbf{z}_{\mathbf{x},0}|\mathbf{z}_{\mathbf{x},1}) = \mathcal{N}(\mathbf{z}_{\mathbf{x},0}| \mathbf{z}_{\mathbf{x},1}/\alpha_{1}-\sigma_{1}/\alpha_{1}\boldsymbol{\epsilon}_{\mathbf{x},0},\sigma_{1}^{2}/\alpha_{1}^{2}\mathbf{I})
\end{equation}

\textbf{Atom Type Features.} For categorical features such as the atom type, the aforementioned integer representation is unnatural and introduces bias. Instead of using integers for these features, we operate directly on a one-hot representation. Suppose $\mathbf{h}$ or $\mathbf{z}_{\mathbf{h},0}$ is an array whose values represent atom types in $\{c_{1},\dots,c_{d}\}$. Then $\mathbf{h}$ is encoded with a one-hot function $\mathbf{h} \gets \mathbf{h}^{\textbf{one-hot}}$ such that $\mathbf{h}_{i,j}^{\textbf{one-hot}} \gets \mathbf{1}_{h_i=c_i}$. and diffusion process over $\mathbf{z}_{\mathbf{h}, t}$ at timestep $t$ and sampling at final timestep are given as:
\begin{align}
    q(\mathbf{z}_{\mathbf{h},t}|\mathbf{z}_{\mathbf{h},0})&=\mathcal{N}(\mathbf{z}_{\mathbf{h},t}|\alpha_{t} \mathbf{h}^{\textbf{one-hot}},\sigma_{t}^{2}\mathbf{I}) \\
    p(\mathbf{z}_{\mathbf{h},0}|\mathbf{z}_{\mathbf{h},1})&=\mathcal{C}(\mathbf{z}_{\mathbf{h},0}|\mathbf{p}),\ \mathbf{p}\propto\int_{\mathbf{1}-\frac{1}{2}}^{\mathbf{1}+\frac{1}{2}} \mathcal{N}(\boldsymbol{u};\mu_{\theta}(\mathbf{z}_{\mathbf{h},1},1),\sigma_{1}^{2})\textrm{d}\boldsymbol{u}
\end{align}
where $\mathbf{p}$ is normalized to sum to one and $\mathcal{C}$ is a categorical distribution.
\par
\textbf{Atom Charge.} Atom charge is the ordinal type of physical quantity, and its sampling process at the final timestep can be formulated by standard practice~\cite{NEURIPS2020_4c5bcfec}:
\begin{align}
    p(\mathbf{z}_{\mathbf{h},0}|\mathbf{z}_{\mathbf{h},1})=\int_{\mathbf{h}-\frac{1}{2}}^{\mathbf{h}+\frac{1}{2}}\mathcal{N}(\boldsymbol{u};\mu_{\theta}(\mathbf{z}_{\mathbf{h},1},1), \sigma_{1}^{2})\textrm{d}\boldsymbol{u}
\end{align}
\textbf{Feature Scaling.} To normalize the features and make them easier to process for the neural network, we add weights to different modalities. The relative scaling has a deeper impact on the model: when the features $\mathbf{h}$ are defined on a smaller scale than the coordinates $\mathbf{x}$, the denoising process tends to first determine rough positions and decide on the atom types only afterward. Empirical knowledge shows that the weights for coordinate, atom type, and atom charge are 1, 0.25, and 0.1, respectively~\cite{pmlr-v162-hoogeboom22a}.
\section{Loss of EMAE is SE(3)-Invariant}
\label{appendix: emae loss}
{\bf Equivariance.} Molecules, typically existing within a three-dimensional physical space, are subject to geometric symmetries, including translations, rotations, and potential reflections. These are collectively referred to as the Euclidean group in 3 dimensions, denoted as~$\mathrm{E(3)}$~\cite{celeghini1991three}. A function $F$ is said to be equivariant to the action of a group $G$ if $T_g \circ F(\mathbf{x}) = F\circ S_g(\mathbf{x})$ for all $g\in G$, where $S_g$, $T_g$ are linear representations related to the group element $g$~\cite{serre1977linear}. We consider the special Euclidean group $\mathrm{SE(3)}$ for geometric graph generation involving translations and rotations. Moreover, the transformations $S_g$ or $T_g$ can be represented by a translation $\mathbf{t}$ and an orthogonal matrix rotation $\mathbf{R}$. For a molecule $\mathcal{G}=\langle\mathbf{x},\mathbf{h}\rangle$, the node features $\mathbf{h}$ are $\mathrm{SE(3)}$-invariant while the coordinates $\mathbf{x}$ are $\mathrm{SE(3)}$-equivariant, which can be expressed as $\mathbf{R}\mathbf{x} + \mathbf{t} = (\mathbf{R}\mathbf{x}_{1} + \mathbf{t},\dots, \mathbf{R}\mathbf{x}_{N} + \mathbf{t})$.

\begin{proof}
\label{proof: emae loss}
\textbf{$\mathcal{L}_{\textrm{\textit{EAAE}}}$ is $SE(3)$-invariance}
\par
Recall the loss function:
\begin{align}
    \mathcal{L}_{\textrm{\textit{EAAE}}} &= \mathbb{E}_{q_{\phi}(\mathbf{f}_{\mathbf{x}}, \mathbf{f}_{\mathbf{h}}|\mathbf{x}^{f},\mathbf{h}^{f})}p_{\vartheta}(\mathbf{x},\mathbf{h}|\mathbf{f}_{\mathbf{x}},\mathbf{f}_{\mathbf{h}})
    -\textrm{KL}[q_{\phi}(\mathbf{f}_{\mathbf{x}}, \mathbf{f}_{\mathbf{h}}|\mathbf{x}^{f},\mathbf{h}^{f})||\prod_{i}^{N}\mathcal{N}({f}_{\mathbf{x},i},{f}_{\mathbf{h}, i}|0,\mathbf{I})] 
\end{align}
Our expected outcome is $\forall\mathbf{R},\ \mathcal{L}_{\textrm{\textit{EAAE}}}(\mathbf{x}, \mathbf{h}, \mathbf{x}^{f}, \mathbf{h}^{f})=\mathcal{L}_{\textrm{\textit{EAAE}}}(\mathbf{R}\mathbf{x}, \mathbf{h}, \mathbf{R}\mathbf{x}^{f}, \mathbf{h}^{f})$. We have:
\begin{equation}
    \begin{aligned}
        &\mathcal{L}_{\textrm{\textit{EAAE}}}(\mathbf{R}\mathbf{x}, \mathbf{h}, \mathbf{R}\mathbf{x}^{f}, \mathbf{h}^{f})\\
    =& \mathbb{E}_{q_{\phi}(\mathbf{f}_{\mathbf{x}}, \mathbf{f}_{\mathbf{h}}|\mathbf{R}\mathbf{x}^{f},\mathbf{h}^{f})}p_{\vartheta}(\mathbf{R}\mathbf{x},\mathbf{h}|\mathbf{f}_{\mathbf{x}},\mathbf{f}_{\mathbf{h}})
    -\textrm{KL}[q_{\phi}(\mathbf{f}_{\mathbf{x}}, \mathbf{f}_{\mathbf{h}}|\mathbf{R}\mathbf{x}^{f},\mathbf{h}^{f})||\prod_{i}^{N}\mathcal{N}({f}_{\mathbf{x},i},{f}_{\mathbf{h}, i}|0,\mathbf{I})]\\
    =&\int_{\mathcal{G}}q_{\phi}(\mathbf{f}_{\mathbf{x}}, \mathbf{f}_{\mathbf{h}}|\mathbf{R}\mathbf{x}^{f},\mathbf{h}^{f})\log p_{\vartheta}(\mathbf{R}\mathbf{x},\mathbf{h}|\mathbf{f}_{\mathbf{x}},\mathbf{f}_{\mathbf{h}})+
        \int_{\mathcal{G}}\log\frac{q_{\phi}(\mathbf{f}_{\mathbf{x}}, \mathbf{f}_{\mathbf{h}}|\mathbf{R}\mathbf{x}^{f},\mathbf{h}^{f})}{\prod_{i}^{N}\mathcal{N}({f}_{\mathbf{x},i},{f}_{\mathbf{h}, i}|0,\mathbf{I})}\\
    =&\int_{\mathcal{G}}q_{\phi}(\mathbf{R}\mathbf{R}^{-1}\mathbf{f}_{\mathbf{x}}, \mathbf{f}_{\mathbf{h}}|\mathbf{R}\mathbf{x}^{f},\mathbf{h}^{f})\log p_{\vartheta}(\mathbf{R}\mathbf{x},\mathbf{h}|\mathbf{R}\mathbf{R}^{-1}\mathbf{f}_{\mathbf{x}},\mathbf{f}_{\mathbf{h}})\\
    &+\int_{\mathcal{G}}\log\frac{q_{\phi}(\mathbf{R}\mathbf{R}^{-1}\mathbf{f}_{\mathbf{x}}, \mathbf{f}_{\mathbf{h}}|\mathbf{R}\mathbf{x}^{f},\mathbf{h}^{f})}{\prod_{i}^{N}\mathcal{N}({f}_{\mathbf{x},i},{f}_{\mathbf{h}, i}|0,\mathbf{I})} && \mathbf{R}\mathbf{R}^{-1}=\mathbf{I} \\
    =&\int_{\mathcal{G}}q_{\phi}(\mathbf{R}^{-1}\mathbf{f}_{\mathbf{x}}, \mathbf{f}_{\mathbf{h}}|\mathbf{x}^{f},\mathbf{h}^{f})\log p_{\vartheta}(\mathbf{x},\mathbf{h}|\mathbf{R}^{-1}\mathbf{f}_{\mathbf{x}},\mathbf{f}_{\mathbf{h}})\\
    &+\int_{\mathcal{G}}\log\frac{q_{\phi}(\mathbf{R}^{-1}\mathbf{f}_{\mathbf{x}}, \mathbf{f}_{\mathbf{h}}|\mathbf{x}^{f},\mathbf{h}^{f})}{\prod_{i}^{N}\mathcal{N}({f}_{\mathbf{x},i},{f}_{\mathbf{h}, i}|0,\mathbf{I})} && SE(3)\ \text{of}\ \mathbf{x},\ \mathbf{f}_{\mathbf{x}},\ \&\ \mathbf{x}^{f} \\
    =&\int_{\mathcal{G}}q_{\phi}(\mathbf{k}, \mathbf{f}_{\mathbf{h}}|\mathbf{x}^{f},\mathbf{h}^{f})\log p_{\vartheta}(\mathbf{x},\mathbf{h}|\mathbf{k},\mathbf{f}_{\mathbf{h}})\cdot|\mathbf{R}|\\
    &+\int_{\mathcal{G}}\log\frac{q_{\phi}(\mathbf{k}, \mathbf{f}_{\mathbf{h}}|\mathbf{x}^{f},\mathbf{h}^{f})}{\prod_{i}^{N}\mathcal{N}({f}_{\mathbf{x},i},{f}_{\mathbf{h}, i}|0,\mathbf{I})} && \text{Let}\ \mathbf{k}=\mathbf{R}^{-1}\mathbf{f}_{\mathbf{x}} \\
    =& \mathbb{E}_{q_{\phi}(\mathbf{k}, \mathbf{f}_{\mathbf{h}}|\mathbf{R}\mathbf{x}^{f},\mathbf{h}^{f})}p_{\vartheta}(\mathbf{x},\mathbf{h}|\mathbf{k},\mathbf{f}_{\mathbf{h}})\\
    &-\textrm{KL}[q_{\phi}(\mathbf{k}, \mathbf{f}_{\mathbf{h}}|\mathbf{x}^{f},\mathbf{h}^{f})||\prod_{i}^{N}\mathcal{N}({f}_{\mathbf{x},i},{f}_{\mathbf{h}, i}|0,\mathbf{I})] && |\mathbf{R}|=1\\
    =& \mathcal{L}_{\textrm{\textit{EAAE}}}(\mathbf{x}^{f}, \mathbf{h}^{f})
    \end{aligned}
\end{equation}
\end{proof}
Given the fragment $\mathcal{G}^{f}$, we subtract the center of gravity from $\mathbf{x}^{f} \in \mathcal{G}^{f}$, and thereby ensure that $\mathcal{E}$ receives isotropic geometric graph, and all together guarantee that the loss of \textit{EAAE} is SE(3)-invariant.
\section{Loss of \textit{GODD} is an SE(3)-Invariant Variational Lower Bound to the Log-likelihood}
\label{appendix: proof}
First, we present the rigorous statement of the Theorem~\ref{thm: loss SE3} here:
\begin{theorem}
Given predefined and valid $\{\alpha_{i}\}_{i=0}^{T}$, $\{\beta_{i}\}_{i=0}^{T}$, and $\{\gamma_{i}\}_{i=0}^{T}$ Let $w(t)$ satisfies:
\begin{align}
    \textrm{1}.&\ \forall t \in [1,\dots,T], w(t) = \frac{\beta_{t}^{2}}{2\gamma_{t}^{2}(1-\beta_{t})(1-\alpha_{t}^{2})}\\
    \textrm{2}.&\ w(0) = -1
\end{align}
Then given the geometric datapoint $\mathcal{G} = \langle\mathbf{x},\mathbf{h}\rangle\in\mathbb{R}^{N\times(3+d)}$ and its subset $\mathcal{G}^{f}\langle\mathbf{x}^{f},\mathbf{h}^{f}\rangle\in\mathbb{R}^{F\times(3+d)}$ the loss $\mathcal{L}$ of the proposed method is expressed as:
\begin{align}
    \mathcal{L} := \mathcal{L}_{\textrm{\textit{EAAE}}} + \mathcal{L}_{\textrm{\textit{DSDM}}}
\end{align}
which satisfies:
\begin{align}
    \textrm{1}.&\ \forall \mathbf{R}~\textrm{and}~\mathbf{t}, ~\mathcal{L}(\mathbf{x}, \mathbf{h},\mathbf{x}^{f},\mathbf{h}^{f})=\mathcal{L}(\mathbf{R}\mathbf{x}+\mathbf{t}, \mathbf{h}, \mathbf{R}\mathbf{x}^{f}+\mathbf{t},\mathbf{h}^{f}) \label{eq: theorem: se3}\\
    \textrm{2}.&\ \mathcal{L}(\mathbf{x}, \mathbf{h}, \mathbf{x}^{f},\mathbf{h}^{f})\geq-\mathbb{E}_{p_{\langle\mathbf{x}, \mathbf{h}\rangle\in\{\mathcal{G}\}},[\mathbf{f}_{\mathbf{x}},\mathbf{f}_{\mathbf{h}}]=\mathcal{E}_{\phi}(\mathcal{G}^{f})}[\log p_{\theta}(\mathbf{z}_{\mathbf{x}},\mathbf{z}_{\mathbf{h}}|\mathbf{f}_{\mathbf{x}},\mathbf{f}_{\mathbf{h}})] \label{eq: theorem: vlb}
\end{align}
And we have $\log p_{\theta}(\mathbf{x}_0, \mathbf{h}_0)$ is the marginal distribution of $\langle\mathbf{x},\mathbf{h}\rangle$ under \textit{GODD}.
\end{theorem}
The theorem proposed herein posits two distinct assertions. Firstly, Equation \ref{eq: theorem: se3} illustrates that the loss function $\mathcal{L}$ is $SE(3)$-invariant, meaning it remains unchanged under any rotational or translational transformations. Secondly, Equation \ref{eq: theorem: vlb} suggests that $\mathcal{L}$ acts as a variational lower bound for the log-likelihood. We provide comprehensive proofs for these assertions separately, commencing with Equation \ref{eq: theorem: vlb}.
\par
\begin{proof}
    \textbf{$\mathcal{L}$ is a variational lower bound of the log-likelihood}
    
    Recall the loss function:
    \begin{align}
        \mathcal{L}(\mathbf{x}, \mathbf{h}, \mathbf{x}^{f},\mathbf{h}^{f}) =& \mathcal{L}_{\textrm{\textit{EAAE}}} + \mathcal{L}_{\textrm{\textit{DSDM}}} \\
        =& \mathbb{E}_{q_{\phi}(\mathbf{f}_{\mathbf{x}}, \mathbf{f}_{\mathbf{h}}|\mathbf{x}^{f},\mathbf{h}^{f})}p_{\vartheta}(\mathbf{x},\mathbf{h}|\mathbf{f}_{\mathbf{x}},\mathbf{f}_{\mathbf{h}})
        -\textrm{KL}[q_{\phi}(\mathbf{f}_{\mathbf{x}}, \mathbf{f}_{\mathbf{h}}|\mathbf{x}^{f},\mathbf{h}^{f})||\prod_{i}^{N}\mathcal{N}({f}_{\mathbf{x},i},{f}_{\mathbf{h}, i}|0,\mathbf{I})] \\
        &+\mathbb{E}_{\mathcal{G},\mathcal{E}_{\phi}(\mathcal{G}^{f}),\epsilon,t}\left[
        \Vert
        \boldsymbol{\epsilon}-\boldsymbol{\epsilon}_{\theta}(\mathbf{x}_{t},\mathbf{h}_{t},\mathbf{f}_{\mathbf{x}},\mathbf{f}_{\mathbf{h}},t)
        \Vert^{2}
        \right]
    \end{align}
    $\mathcal{L}_{\textrm{\textit{EAAE}}}$ is a standard variational autoencoder and has been proven to be a variational lower bound of the log-likelihood~\cite{DBLP:journals/corr/KingmaW13}. For simplicity, we denote $\mathbf{z}_{\mathbf{x},t}, \mathbf{z}_{\mathbf{h},t}$ as $\mathbf{z}_{t}$, and $\mathbf{f}_{\mathbf{x}}, \mathbf{f}_{\mathbf{h}}$ as $\mathbf{f}$, then we expect $\mathcal{L}_{\textrm{\textit{DSDM}}}$ has:
    \begin{align}
        \log p_{\theta}(\mathbf{z}|\mathbf{f})\geq \text{KL}[q(\mathbf{z}_{1:T}|\mathbf{z}_{0})\Vert p_{\theta}(\mathbf{z}|\mathbf{f})]
    \end{align}
    \begin{equation}
        \begin{aligned}
            \log p_{\theta}(\mathbf{z}|\mathbf{f})
            \geq&\mathbb{E}_{q(\mathbf{z}_{1:T}|\mathbf{z}_{0})}\left[\log\frac{p_{\theta}(\mathbf{z}_{0:T}|\mathbf{f})}{q(\mathbf{z}_{1:T}|\mathbf{z}_{0})}\right]\\
            =&\mathbb{E}_{q(\mathbf{z}_{1:T}|\mathbf{z}_{0})}\left[\log\frac{p(\mathbf{z}_{T})p_{\theta}(\mathbf{z}_{0}|\mathbf{z}_{1},\mathbf{f})\prod_{t=2}^{T}p_{\theta}(\mathbf{z}_{t-1}|\mathbf{z}_{t}, \mathbf{f})}{q(\mathbf{z}_{1}|\mathbf{z}_{0})\prod_{t=2}^{T}q(\mathbf{z}_{t}|\mathbf{z}_{t-1})}\right]\\
            =&\mathbb{E}_{q(\mathbf{z}_{1:T}|\mathbf{z}_{0})}\left[\log\frac{p(\mathbf{z}_{T})p_{\theta}(\mathbf{z}_{0}|\mathbf{z}_{1},\mathbf{f})}{q(\mathbf{z}_{1}|\mathbf{z}_{0})} + \log
            \prod_{t=2}^{T}\frac{p_{\theta}(\mathbf{z}_{t-1}|\mathbf{z}_{t}, \mathbf{f})}{q(\mathbf{z}_{t}|\mathbf{z}_{t-1})}\right]\\
            =&\mathbb{E}_{q(\mathbf{z}_{1:T}|\mathbf{z}_{0})}\left[\log\frac{p(\mathbf{z}_{T})p_{\theta}(\mathbf{z}_{0}|\mathbf{z}_{1},\mathbf{f})}{q(\mathbf{z}_{1}|\mathbf{z}_{0})} + \log
            \prod_{t=2}^{T}\frac{p_{\theta}(\mathbf{z}_{t-1}|\mathbf{z}_{t}, \mathbf{f})}{\frac{q(\mathbf{z}_{t-1}|\mathbf{z}_{t},\mathbf{z}_{0})q(\mathbf{z}_{t}|\mathbf{z}_{0})}{q(\mathbf{z}_{t-1}|\mathbf{z}_{0})}}\right]\\
            =&\mathbb{E}_{q(\mathbf{z}_{1:T}|\mathbf{z}_{0})}\left[\log\frac{p(\mathbf{z}_{T})p_{\theta}(\mathbf{z}_{0}|\mathbf{z}_{1},\mathbf{f})}{q(\mathbf{z}_{T}|\mathbf{z}_{0})} + \sum_{t=2}^{T}\log
            \frac{p_{\theta}(\mathbf{z}_{t-1}|\mathbf{z}_{t}, \mathbf{f})}{q(\mathbf{z}_{t-1}|\mathbf{z}_{t},\mathbf{z}_{0})}\right]\\
            =&\mathbb{E}_{q(\mathbf{z}_{1}|\mathbf{z}_{0})}[p_{\theta}(\mathbf{z}_{0}|\mathbf{z}_{1},\mathbf{f})]+\mathbb{E}_{q(\mathbf{z}_{T}|\mathbf{z}_{0})}\left[\log\frac{p(\mathbf{z}_{T})}{q(\mathbf{z}_{T}|\mathbf{z}_{0})}\right] \\
            & + \sum_{t=2}^{T}\mathbb{E}_{q(\mathbf{z}_{t},\mathbf{z}_{t-1}|\mathbf{z}_{0})}\left[\log
            \frac{p_{\theta}(\mathbf{z}_{t-1}|\mathbf{z}_{t}, \mathbf{f})}{q(\mathbf{z}_{t-1}|\mathbf{z}_{t},\mathbf{z}_{0})}\right]\\
            =&\mathbb{E}_{q(\mathbf{z}_{1}|\mathbf{z}_{0})}[p_{\theta}(\mathbf{z}_{0}|\mathbf{z}_{1},\mathbf{f})]-\text{KL}[q(\mathbf{z}_{T}|\mathbf{z}_{0})\Vert p(\mathbf{z}_{T})]\\
            &-\sum_{t=2}^{T}\mathbb{E}_{q(\mathbf{z}_{t}|\mathbf{z}_{0})}[\text{KL}[q(\mathbf{z}_{t-1}|\mathbf{z}_{t},\mathbf{z}_{0})\Vert p_{\theta}(\mathbf{z}_{t-1}|\mathbf{z}_{t}, \mathbf{f})]]
        \end{aligned}
    \end{equation}
    where we denote $\text{KL}[q(\mathbf{z}_{t-1}|\mathbf{z}_{t},\mathbf{z}_{0})\Vert p_{\theta}(\mathbf{z}_{t-1}|\mathbf{z}_{t}, \mathbf{f})]$ as $\mathcal{L}_{\textrm{\textit{DSDM}}, t-1}$, then we have:
    \begin{equation}
        \begin{aligned}
            \mathcal{L}_{\textrm{\textit{DSDM}}, t-1} = \mathbb{E}_{\boldsymbol{\epsilon}\sim\mathcal{N}(0,\mathbf{I})}\left[ \frac{\beta_{t}^{2}}{2\gamma_{t}^{2}(1-\beta_{t})(1-\alpha_{t}^{2})}\Vert\boldsymbol{\epsilon}-\boldsymbol{\epsilon}_{\theta}(\mathbf{z}_{t},\mathbf{f},t)\Vert_{2}^{2} \right]
        \end{aligned}
    \end{equation}
    which gives us the weights of $w(t)$ for $t = 1,\dots,T$.
    \par
    For term $\mathbb{E}_{q(\mathbf{z}_{1}|\mathbf{z}_{0})}[p_{\theta}(\mathbf{z}_{0}|\mathbf{z}_{1},\mathbf{f})]$, we denote as $\mathcal{L}_{\textrm{\textit{DSDM}}, 0}$. With sampling at the final timestep for different modality features and a normalization constant $Z$, we have:
    \begin{equation}
        \begin{aligned}
            \mathcal{L}_{\textrm{\textit{DSDM}}, 0} = \mathbb{E}_{\boldsymbol{\epsilon}\sim\mathcal{N}(0,\mathbf{I})}\left[ 
            \log Z^{-1} - \frac{1}{2}\Vert\boldsymbol{\epsilon}-\boldsymbol{\epsilon}_{\theta}(\mathbf{z},\mathbf{f},1) \Vert^{2}
            \right]
        \end{aligned}
    \end{equation}
    \par
    Since $\mathbf{z}_{T}\sim\mathcal{N}(0,\mathbf{I})$, we have:
    \begin{equation}
        \mathcal{L}_{\textrm{\textit{DSDM}}, T} = \text{KL}[q(\mathbf{z}_{T}|\mathbf{z}_{0})\Vert p(\mathbf{z}_{T})]=0
    \end{equation}
    Therefore, we have:
    \begin{equation}
        \begin{aligned}
            \mathbb{E}_{p_{\langle\mathbf{x}, \mathbf{h}\rangle\in\{\mathcal{G}\}},[\mathbf{f}_{\mathbf{x}},\mathbf{f}_{\mathbf{h}}]=\mathcal{E}_{\phi}(\mathcal{G}^{f})}[\log p_{\theta}(\mathbf{z}|\mathbf{f})] \geq -\sum_{t=2}^{T}\mathcal{L}_{\textrm{\textit{DSDM}}, t-1}-\mathcal{L}_{\textrm{\textit{DSDM}}, 0} = -\mathcal{L}_{\textrm{\textit{DSDM}}}
        \end{aligned}
    \end{equation}
\end{proof}
\par
We then prove Equation \ref{eq: theorem: se3}:
\begin{proof}
\textbf{$\mathcal{L}$ is $SE(3)$-invariance}
\par
Our expected outcome is $\forall\mathbf{R},\ \mathcal{L}(\mathbf{x}, \mathbf{h}, \mathbf{x}^{f},\mathbf{h}^{f})=\mathcal{L}(\mathbf{R}\mathbf{x}, \mathbf{h},\mathbf{R}\mathbf{x}^{f},\mathbf{h}^{f})$, and $\forall\mathbf{R}$, $\mathcal{L}_{\textrm{\textit{EAAE}}}(\mathbf{x}, \mathbf{h}, \mathbf{x}^{f}, \mathbf{h}^{f})=\mathcal{L}_{\textrm{\textit{EAAE}}}(\mathbf{R}\mathbf{x}, \mathbf{h},\mathbf{R}\mathbf{x}^{f},\mathbf{h}^{f})$ is ensured in Proof. \ref{proof: emae loss}. For $\mathcal{L}_{\textrm{\textit{DSDM}}}$, we expect $\forall\mathbf{R}, \mathcal{L}_{\textrm{\textit{DSDM}}}(\mathbf{R}\mathbf{z}_{\mathbf{x},0},\mathbf{z}_{\mathbf{h},0}, \mathbf{f})=\mathcal{L}_{\textrm{\textit{DSDM}}}(\mathbf{z}_{\mathbf{x},0},\mathbf{z}_{\mathbf{h},0}, \mathbf{f})$ we have:
\begin{equation}
    \begin{aligned}
    \nonumber
        & \mathcal{L}_{\textrm{\textit{DSDM}}}(\mathbf{R}\mathbf{z}_{\mathbf{x},0},\mathbf{z}_{\mathbf{h},0})\\
        = & \mathbb{E}_{\mathcal{G}, \mathcal{E}_{\phi}}\left[\sum_{t=2}^{T}\mathbb{E}_{q(\mathbf{z}_{t}|\mathbf{R}\mathbf{z}_{0})}[\text{KL}[q(\mathbf{z}_{t-1}|\mathbf{z}_{t},\mathbf{R}\mathbf{z}_{0})\Vert p_{\theta}(\mathbf{z}_{t-1}|\mathbf{z}_{t},\mathbf{f})]]-\mathbb{E}_{q(\mathbf{z}_{1}|\mathbf{R}\mathbf{z}_{0})}[p_{\theta}(\mathbf{R}\mathbf{z}_{0}|\mathbf{z}_{1},\mathbf{f})] \right]\\
    \end{aligned}
\end{equation}
\begin{equation}
    \begin{aligned}
        = & \int_{\mathcal{G}}\left[ \sum_{t=2}^{T}\log\frac{q(\mathbf{z}_{t-1}|q(\mathbf{z}_{t},\mathbf{R}\mathbf{z}_{0})}{p_{\theta}(\mathbf{z}_{t-1}|\mathbf{z}_{t},\mathbf{f})}-\log p_{\theta}(\mathbf{R}\mathbf{z}_{0}|\mathbf{z}_{1},\mathbf{f}) \right] \\
        = & \int_{\mathcal{G}}\left[ \sum_{t=2}^{T}\log\frac{q(\mathbf{R}\mathbf{R}^{-1}\mathbf{z}_{t-1}|q(\mathbf{R}\mathbf{R}^{-1}\mathbf{z}_{t},\mathbf{R}\mathbf{z}_{0})}{\mathbf{R}\mathbf{R}^{-1}p_{\theta}(\mathbf{z}_{t-1}|\mathbf{R}\mathbf{R}^{-1}\mathbf{z}_{t},\mathbf{f})}-\log p_{\theta}(\mathbf{R}\mathbf{z}_{0}|\mathbf{R}\mathbf{R}^{-1}\mathbf{z}_{1},\mathbf{f}) \right] && \mathbf{R}\mathbf{R}^{-1}=\mathbf{I} \\
        = & \int_{\mathcal{G}}\left[ \sum_{t=2}^{T}\log\frac{q(\mathbf{R}^{-1}\mathbf{z}_{t-1}|q(\mathbf{R}^{-1}\mathbf{z}_{t},\mathbf{z}_{0})}{\mathbf{R}^{-1}p_{\theta}(\mathbf{z}_{t-1}|\mathbf{R}^{-1}\mathbf{z}_{t},\mathbf{f})}-\log p_{\theta}(\mathbf{z}_{0}|\mathbf{R}^{-1}\mathbf{z}_{1},\mathbf{f}) \right] && SE(3)\ \text{of} \ \mathbf{f}_{\mathbf{x}}\ \&\ \mathbf{z}_{t} \\
        = & \mathbb{E}_{\mathcal{G}, \mathcal{E}_{\phi}} \left[ \sum_{t=2}^{T}\log\frac{q(\mathbf{j}_{t-1}|q(\mathbf{j}_{t},\mathbf{z}_{0})}{\mathbf{R}^{-1}p_{\theta}(\mathbf{z}_{t-1}|\mathbf{j}_{t},\mathbf{f})}-\log p_{\theta}(\mathbf{z}_{0}|\mathbf{j}_{1},\mathbf{f}) \right] && \text{Let}\ \mathbf{j}_{t}=\mathbf{R}^{-1}\mathbf{z}_{t} \\
        = & \mathcal{L}_{\text{\textit{DSDM}}}(\mathbf{z}_{\mathbf{x},0},\mathbf{z}_{\mathbf{h},0}, \mathbf{f})
    \end{aligned}
\end{equation}
\end{proof}

\section{Training Details}
\label{appendix: parameters}
\textbf{Parameters}
\begin{enumerate}
    \item Optimizer: Adam~\cite{KingBa15} optimizer is used with a constant learning rate of $10^{-4}$ as our default training configuration.
    \item Batch size: 64.
    \item EGNN in \textit{DSDM}: 9 layers and 256 hidden features for QM9, 4 layers and 256 hidden features for GEOM-DRUG.
    \item EGNN in \textit{EAAE}: 1 layer and 256 hidden features for the encoder for QM9 and GEOM-DRUG, 9 layers and 4 layers with 256 hidden features for the decoder for QM9 and GEOM-DRUG, respectively.
    \item Latent dimension of $\mathbf{f}_{\mathbf{x}}, \mathbf{f}_{\mathbf{h}}$: latent dimension is 3 and 1 for $\mathbf{f}_{\mathbf{x}}$ and $\mathbf{f}_{\mathbf{h}}$, respectively.
    \item Epoch: 3000 for QM9 and 10 for GEOM-DRUG.
\end{enumerate}
\textbf{Training}
\begin{enumerate}
    \item GPU: NVIDIA GeForce RTX 3090
    \item CPU: Intel(R) Xeon(R) Platinum 8338C CPU
    \item Memory: 512 GB
    \item Time: Around 7 days for QM9 and 20 days for GEOM-DRUG.
\end{enumerate}

\textbf{Specific Parameters}
1. On QM9, we train \textit{DSDM} with 9 layers and 256 hidden features with a batch size 64;
2. On GEOM-DRUG, we train \textit{DSDM} with 4 layers and 256 hidden features, with batch size 64;

\section{Algorithms}
\label{appendix: algorithm}
This section contains two main algorithms of the proposed \textit{GODD}. Algorithm~\ref{alg: training} presents the pseudo-code for training \textit{GODD} on the in distributional training data set $\{\mathcal{G}_{I}\}$ and corresponding fragment set $\{\mathcal{G}^{f}_{I}\}$. Algorithm~\ref{alg: sampling} presents the process of OOD molecule generation using the ODD scaffold/ring $\mathcal{G}^{f}_{O}$.
\begin{algorithm}[htb]
   \caption{Training \textit{GODD}}
   \label{alg: training}
    \begin{algorithmic}[1]
       \STATE {\bfseries Input:} in-distribution geometric data point $\mathcal{G}_{I}=\langle\mathbf{x},\mathbf{h}\rangle$, corresponding fragment $\mathcal{G}^{f}_{I}$, asymmetric encoder $\mathcal{E}_{\phi}$ and decoder $\mathcal{D}_{\vartheta}$, denoising network $\boldsymbol{\epsilon}_{\theta}$;
       \STATE \textbf{\textit{EAAE}:}
       \STATE $\boldsymbol{\mu}_{\boldsymbol{x}}, \mathbf{\mu}_{\mathbf{h}} \gets \mathcal{E}_{\phi}(\mathbf{x}^{f},\mathbf{h}^{f} )$ \hfill \textit{//~Encode}
       \STATE $\langle\boldsymbol{\epsilon}_{\mathbf{x}},\boldsymbol{\epsilon}_{\mathbf{h}}\rangle\sim\mathcal{N}(\mathbf{0},\mathbf{I})$  \hfill \textit{//~Sample Noise for \textit{EAAE}}
       \STATE $\boldsymbol{\epsilon}_{\mathbf{x}} \gets \boldsymbol{\epsilon}_{\mathbf{x}} - \mathbf{G}(\boldsymbol{\epsilon}_{\mathbf{x}})$ \hfill\textit{//~Subtract Center of Gravity}
       \STATE $\mathbf{f}_{\mathbf{x}},\mathbf{f}_{\mathbf{h}}\gets \mu+\langle\boldsymbol{\epsilon}_{\mathbf{x}},\boldsymbol{\epsilon}_{\mathbf{h}}\rangle\odot\sigma_{0}$\hfill\textit{//~Reparameterization}
       \STATE \textbf{\textit{DSDM}:}
       \STATE $t\sim\mathcal{U}(0, T)$\hfill \textit{//~Sample Timestep}
       \STATE $\langle\boldsymbol{\epsilon}_{\mathbf{x}},\boldsymbol{\epsilon}_{\mathbf{h}}\rangle\sim\mathcal{N}(\mathbf{0},\mathbf{I})$  \hfill \textit{//~Sample Noise for \textit{DSDM}}
       \STATE $\boldsymbol{\epsilon}_{\mathbf{x}} \gets \boldsymbol{\epsilon}_{\mathbf{x}} - \mathbf{G}(\boldsymbol{\epsilon}_{\mathbf{x}})$ \hfill\textit{//~Subtract Center of Gravity}
       \STATE $\mathbf{z}_{\mathbf{x},t},\mathbf{z}_{\mathbf{h},t} \gets \alpha_{t}[\mathbf{x}, \mathbf{h}] + \sigma_{t}\boldsymbol{\epsilon}$ \hfill\textit{//~Diffuse}
       \STATE $\hat{\mathbf{x}},\hat{\mathbf{h}}\gets\mathcal{D}_{\vartheta}(\mathbf{f}_{\mathbf{x}},\mathbf{f}_{\mathbf{h}})$\hfill\textit{//~Decode}
       \STATE \textbf{Optimization}
       \STATE $\mathcal{L}_{\textrm{\textit{EAAE}}}\gets\mathcal{L}([\hat{\mathbf{x}},\hat{\mathbf{h}}],[{\mathbf{x}},{\mathbf{h}}])+\textrm{KL}$ \hfill\textit{//~$\mathcal{L}$ for \textit{EAAE}}
       \STATE $\mathcal{L}_{\textrm{\textit{DSDM}}} \gets \Vert\boldsymbol{\epsilon}-\boldsymbol{\epsilon}_{\theta}(\mathbf{z}_{\mathbf{x},t},\mathbf{z}_{\mathbf{h},t}, t, \mathbf{f}_{\mathbf{x}},\mathbf{f}_{\mathbf{h}})\Vert^{2} $
       \hfill\textit{//~$\mathcal{L}$ for \textit{DSDM}}
       \STATE $\mathcal{L}_{\textrm{\textit{GODD}}}\gets\mathcal{L}_{\textrm{\textit{EAAE}}}+\mathcal{L}_{\textrm{\textit{DSDM}}}$\hfill\textit{//~Total Loss}
       \STATE $\phi,\vartheta,\theta\gets\textrm{optimizer}(\mathcal{L}_{\textrm{\textit{GODD}}},\phi,\vartheta,\theta)$\hfill\textit{//~Optimize}
       \STATE {\bfseries return} $\phi,\theta$
    \end{algorithmic}
\end{algorithm}

\begin{algorithm}[htb]
   \caption{Adaptive Sampling of \textit{GODD}}
   \label{alg: sampling}
    \begin{algorithmic}[1]
       \STATE {\bfseries Input:} OOD fragment $\mathcal{G}^{f}_{O}=\langle\mathbf{x}^{f}_{O},\mathbf{h}^{f}_{O}\rangle$, encoder $\mathcal{E}_{\phi}$, denoising network $\boldsymbol{\epsilon}_{\theta}$;
        \STATE $\boldsymbol{\mu}_{\boldsymbol{x}}, \mathbf{\mu}_{\mathbf{h}} \gets \mathcal{E}_{\phi}(\mathbf{x}^{f}_{O},\mathbf{h}^{f}_{O})$ \hfill \textit{//~Encode}
       \STATE $\langle\boldsymbol{\epsilon}_{\mathbf{x}},\boldsymbol{\epsilon}_{\mathbf{h}}\rangle\sim\mathcal{N}(\mathbf{0},\mathbf{I})$  \hfill \textit{//~Sample Noise for \textit{EAAE}}
       \STATE $\boldsymbol{\epsilon}_{\mathbf{x}} \gets \boldsymbol{\epsilon}_{\mathbf{x}} - \mathbf{G}(\boldsymbol{\epsilon}_{\mathbf{x}})$ \hfill\textit{//~Subtract Center of Gravity}
       \STATE $\mathbf{f}_{\mathbf{x}},\mathbf{f}_{\mathbf{h}}\gets \mu+\langle\boldsymbol{\epsilon}_{\mathbf{x}},\boldsymbol{\epsilon}_{\mathbf{h}}\rangle\odot\sigma_{0}$\hfill\textit{//~Target Condition}

       \STATE $\langle\mathbf{z}_{\mathbf{x},T},\mathbf{z}_{\mathbf{h},T}\rangle\sim\mathcal{N}(\mathbf{0},\mathbf{I})$  \hfill \textit{//~Sample Noise for Generation}
       \FOR{$t$ \textbf{in} $T, T-1,\dots, 1$}
            \STATE $\langle\boldsymbol{\epsilon}_{\mathbf{x}},\boldsymbol{\epsilon}_{\mathbf{h}}\rangle\sim\mathcal{N}(\mathbf{0},\mathbf{I})$  \hfill \textit{//~Denoising}
            \STATE $\boldsymbol{\epsilon}_{\mathbf{x}} \gets \boldsymbol{\epsilon}_{\mathbf{x}} - \mathbf{G}(\boldsymbol{\epsilon}_{\mathbf{x}})$ \hfill\textit{//~Subtract Center of Gravity}
            \STATE $\mathbf{z}_{\mathbf{x},t-1},\mathbf{z}_{\mathbf{h},t-1} \gets \frac{1}{\sqrt{1-\beta_{t}}}(\langle\mathbf{z}_{\mathbf{x},t},\mathbf{z}_{\mathbf{h},t}\rangle-\frac{\beta_{t}}{\sqrt{1-\alpha_{t}^{2}}}\boldsymbol{\epsilon}_{\theta}(\mathbf{z}_{\mathbf{x},t},\mathbf{z}_{\mathbf{h},t},t,\mathbf{f}_{\mathbf{x}},\mathbf{f}_{\mathbf{h}})) + \rho_{t}\boldsymbol{\epsilon}$ 
       \ENDFOR
       \STATE$\mathbf{x},\mathbf{h}\gets p(\mathbf{z}_{\mathbf{x},0},\mathbf{z}_{\mathbf{h},0}|\mathbf{z}_{\mathbf{x},1},\mathbf{z}_{\mathbf{h},1},\mathbf{f}_{\mathbf{x}},\mathbf{f}_{\mathbf{h}})$
       \STATE {\bfseries return} $\langle\mathbf{x},\mathbf{h}\rangle$
    \end{algorithmic}
\end{algorithm}

\section{\emph{QM9} Dataset}
\label{appendix: qm9}
QM9~\cite{QM9} is a comprehensive dataset that provides geometric, energetic, electronic, and thermodynamic properties for a subset of the GDB-17 database~\cite{doi:10.1021/ci300415d}, comprising 134 thousand stable organic molecules with up to nine heavy atoms.
\subsection{Scaffold Split QM9}
We utilized the open-source software, RDkit~\cite{landrum2016rdkit}, to examine the scaffold and ring of each molecule. QM9 dataset~\footnote{\url{https://springernature.figshare.com/ndownloader/files/3195389}} comprises a total of 130,831 molecules, encompassing 15,661 unique scaffolds. Molecules lacking a scaffold were denoted as ‘-’ and included in the total scaffold count. The dataset was divided based on scaffold frequency. Specifically, the in-distribution subset contained 100,000 molecules and 1,054 scaffolds. The OOD I subset included 15,000 molecules and 2,532 scaffolds, while the OOD II subset consisted of 15,831 molecules and 12,075 scaffolds.
\par
\begin{figure}[ht]
\begin{center}
    \subfigure[The number of molecules and scaffolds in distribution, OOD I, and OOD II of the Scaffold-Split QM9 data set.]{
        \label{fig: scaffold qm9-1}
        \includegraphics[width=0.96\columnwidth]{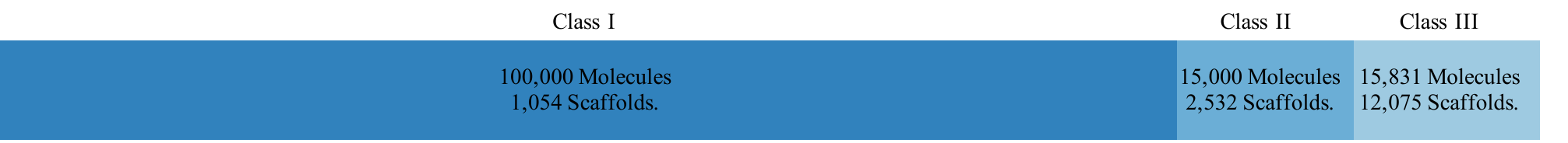}
    }\\
    \subfigure[Scaffold Logarithmic Histogram of Scaffold-Split QM9]{
    \label{fig: scaffold qm9-2}
        \includegraphics[width=\columnwidth]{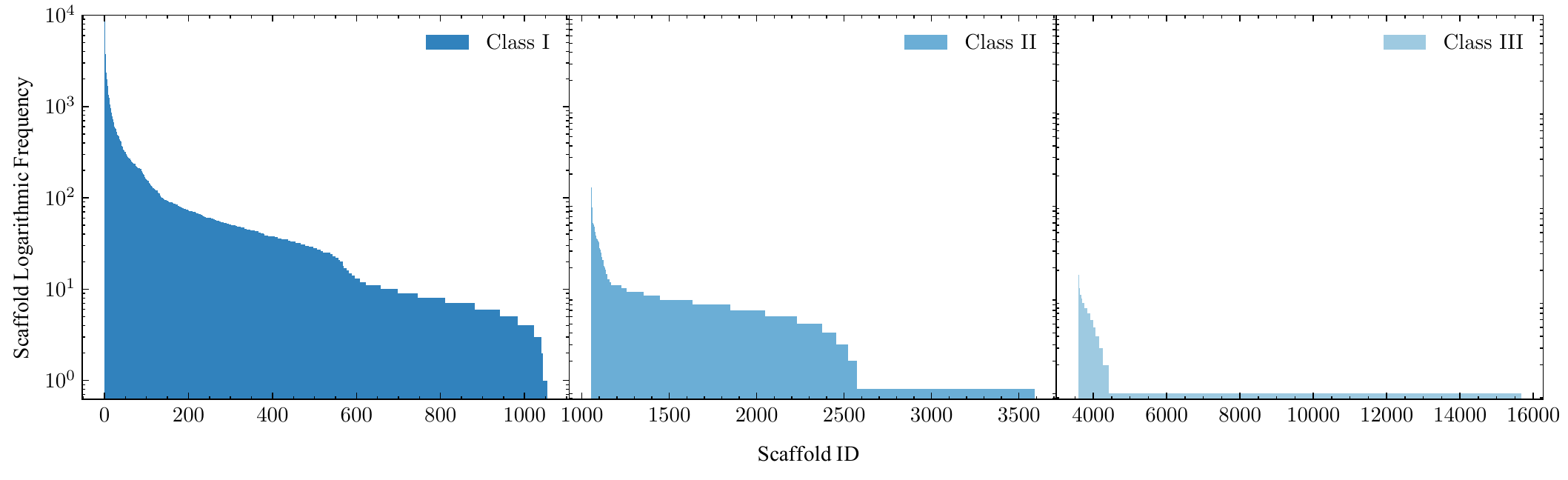}
    }
\caption{Scaffold-Split QM9}
\label{fig: scaffold qm9}
\end{center}
\end{figure}
\begin{figure}[ht]
\begin{center}
\centerline{
    \includegraphics[width=\columnwidth]{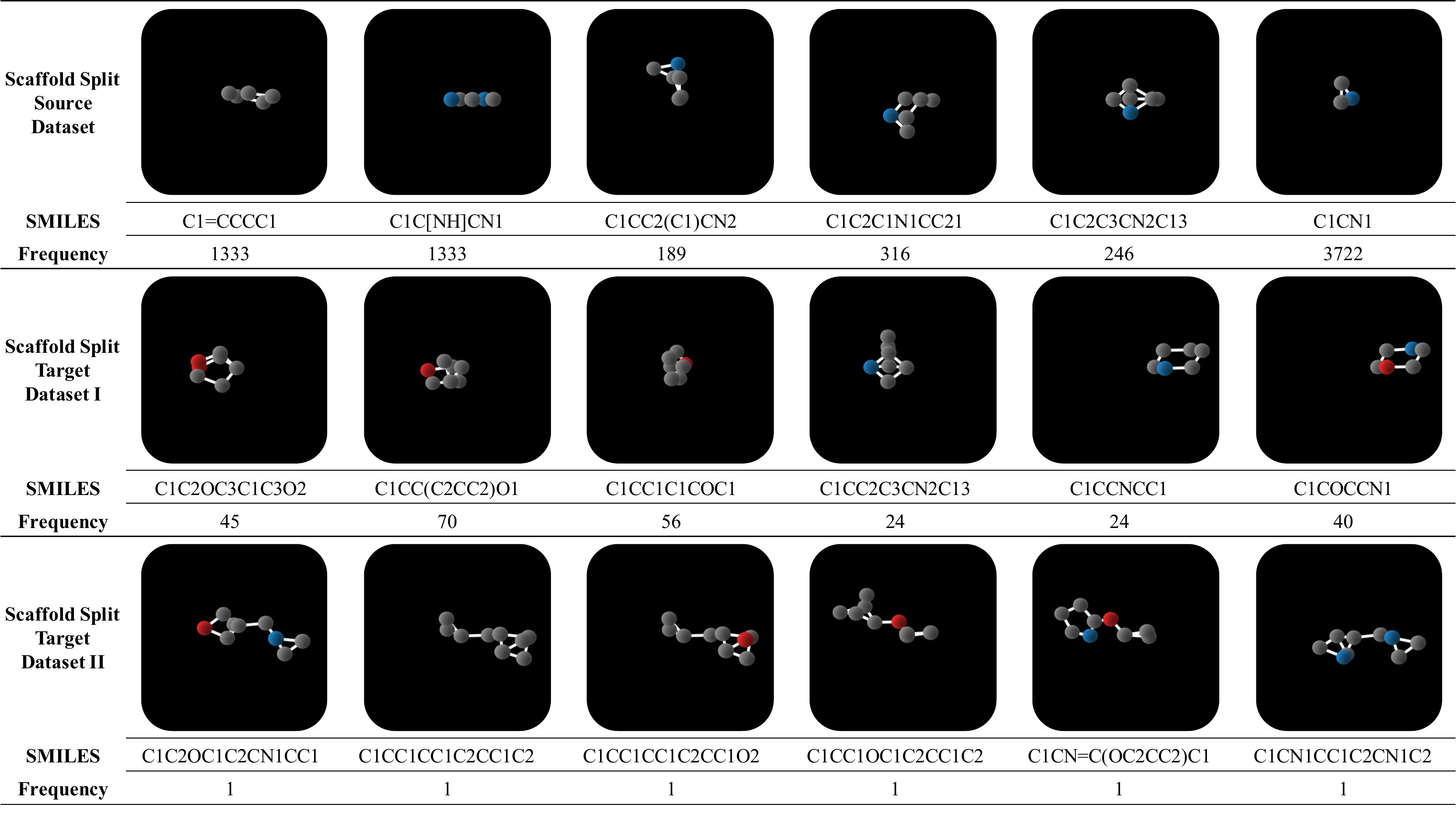}
}
\caption{Scaffold Examples of QM9 Split by Scaffolds.}
\label{fig: scaffold qm9 examples}
\end{center}
\end{figure}

Figure~\ref{fig: scaffold qm9-1} presents the division's schematic diagram. Figure~\ref{fig: scaffold qm9-2} displays the logarithmic histogram of the scaffolds in each dataset segment. It is evident that the in-distribution dataset contains the most frequent scaffolds, primarily concentrated above 100. The frequency of scaffolds in the OOD I dataset ranges between 10 and 100. In contrast, the scaffolds in the OOD II dataset are primarily concentrated within 10, with most appearing only once. Figures, SMILES, and frequencies of some example scaffolds in each sub-dataset are given in Figure~\ref{fig: scaffold qm9 examples}.

\subsection{Ring Number Split QM9}
\begin{figure}[ht]
\begin{center}
\centerline{
    \includegraphics[width=\columnwidth]{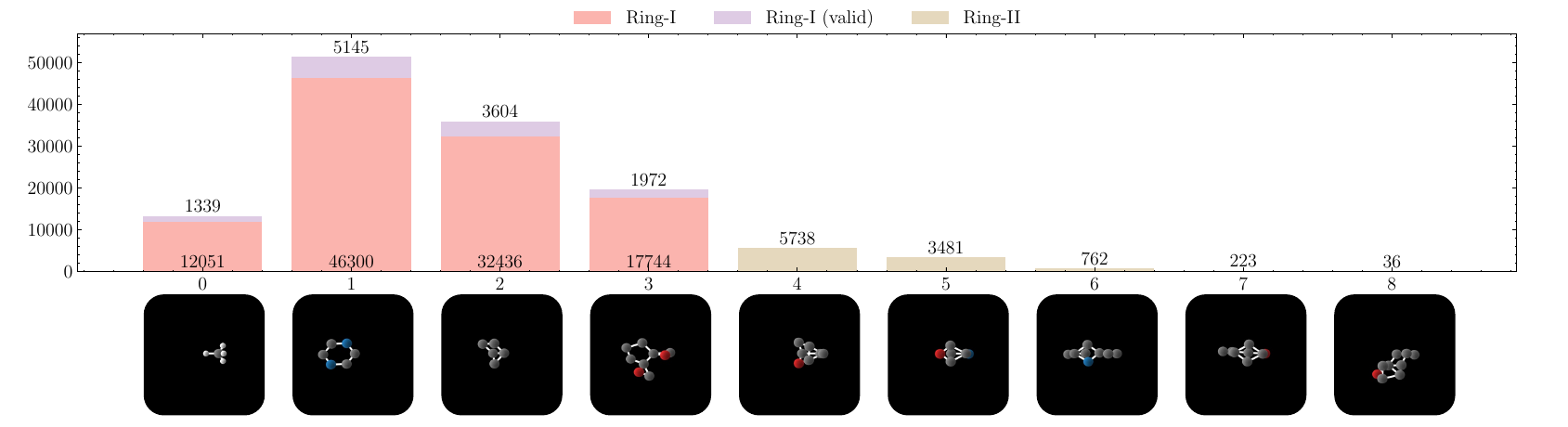}
}
\caption{Ring Examples of QM9 Split by Ring Number.}
\label{fig: ring qm9}
\end{center}
\end{figure}

The QM9 dataset could categorize molecules into nine groups based on the number of rings, ranging from 0 to 8. As the number of rings increases, the quantity of molecules correspondingly decreases. We partition the QM9 dataset into two subsets based on ring count. The in-distribution dataset comprises acyclic molecules and those with 1 to 3 rings, while the OOD dataset includes molecules with 4 to 8 rings. Figure~\ref{fig: ring qm9} presents a schematic diagram illustrating example molecules with 0 to 8 rings.

\section{\textit{GEOM-DRUG} Dataset}
\label{appendix: drug}
\textit{GEOM-DRUG} (Geometric Ensemble Of Molecules) dataset~\cite{geomdrug} encompasses around 450,000 molecules, each with an average of 44.2 atoms and a maximum of 181 atoms\footnote{\url{https://dataverse.harvard.edu/file.xhtml?fileId=4360331&version=2.0}}.
\par
\subsection{Ring Number Split GEOM-DRUG}
\begin{figure}[ht]
\begin{center}
\centerline{
    \includegraphics[width=\columnwidth]{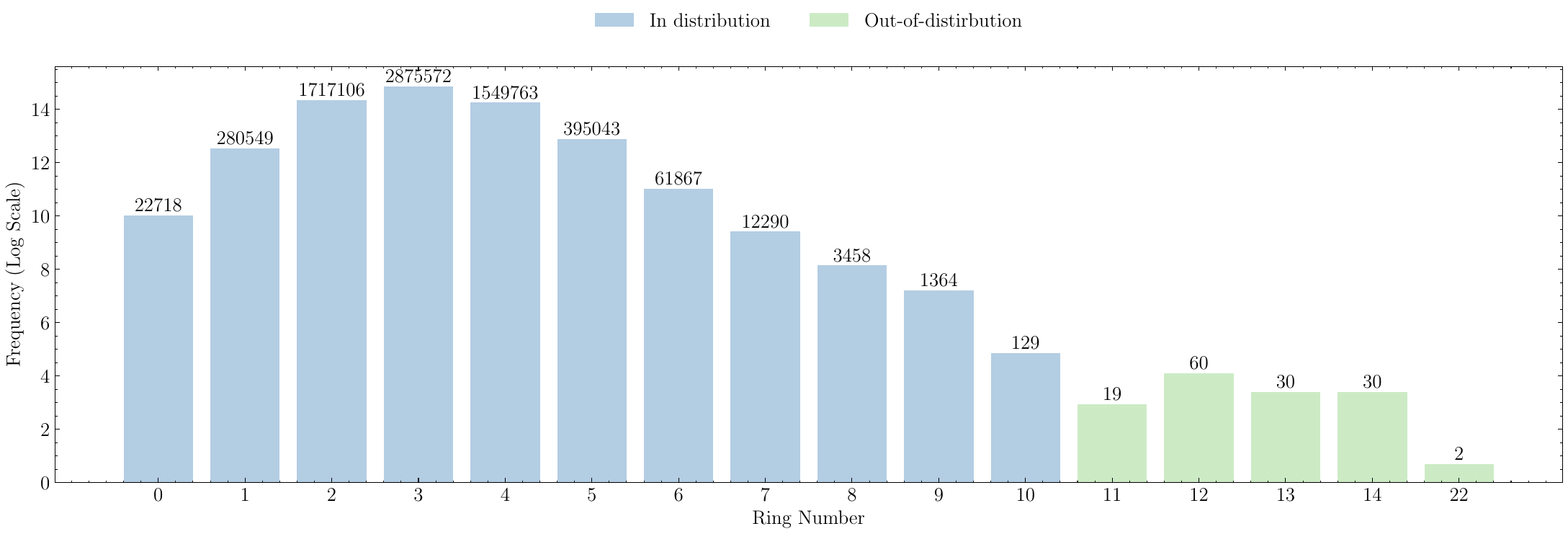}
}
\caption{Ring Distribution of GEOM-DRUG dataset.}
\label{fig: ring drug}
\end{center}
\end{figure}

The GEOM-DRUG dataset classifies molecules into sixteen categories based on the number of rings, ranging from 0 to 14 and 22. As the ring count increases, the number of molecules correspondingly decreases. The dataset is partitioned into two subsets according to ring count: the in-distributional dataset, which includes molecules with 0 to 10 rings and a count exceeding 100, and four OOD datasets, which comprises molecules with 11 to 14 and 22 rings. Figure~\ref{fig: ring drug} provides a schematic representation of the molecule distribution by ring number.
\section{\textit{GEOM-LINKER} Dataset}
\label{appendix: linker}
The GEOM-LINKER dataset for linker design is constructed by~\cite{linker-design-NMI} based on GEOM-DRUG. The authors decomposed the molecule into three or more fragments with one or two linkers connecting them. The dataset contains 41,907 molecules and 285,140 fragments, and the original dataset is randomly split into train (282,602 examples), validation (1,250 examples), and test (1,288 examples) sets. Atom types considered for this dataset are C, O, N, F, S, Cl, Br, I, and P.
\par
\begin{figure}[ht]
\begin{center}
\centerline{
    \includegraphics[width=\columnwidth]{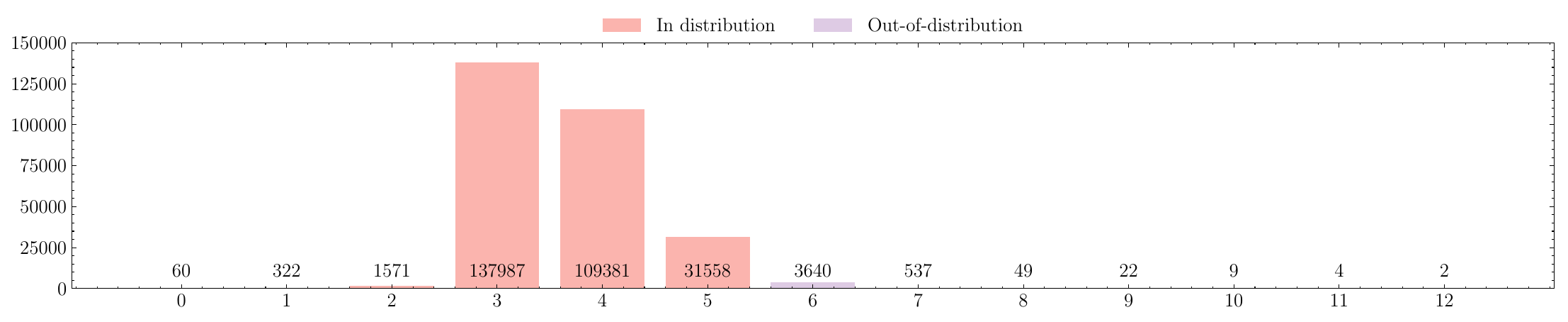}
}
\caption{Ring Distribution of GEOM-LINKER dataset.}
\label{fig: linker hist}
\end{center}
\end{figure}
We present the distribution of molecules in GEOM-LINKER according to the number of rings in Figure~\ref{fig: linker hist}. The diagram illustrates the molecules with 3 to 5 rings are the majority and molecules with 8 to 12 rings exhibit data sparsity in the whole dataset. Thereby, we split the dataset according to the ring numbers into in-distribution (0-5 rings, 280,879 samples) and OOD (6-12 rings, 4,263 samples). 

\section{Full Results of OOD Ring-Structure Molecule Generation}
\label{appendix: ring results}
We present the detailed quantitative evaluation results of ring adaptive molecule generation tasks in Tables~\ref{tab: full ring results-1}, \ref{tab: full ring results-2}, \ref{tab: full ring results-3}, and~\ref{tab: full ring results-4}. The results show that the proposed method has dominant performance in all metrics, including ring number proportion, validity, novelty, and success rate.
\par
It is significant to note that the entire QM9 dataset comprises only 36 eight-ring molecules. When the proposed algorithm utilizes the ring structures of these 36 8-ring molecules as input, the target validity reaches an impressive 72.2\%, and the novelty is as high as 80.9\%.
\begin{table}[htbp]
  \centering
  \caption{Results of molecule proportion in terms of ring-number (P). The \textbf{best} results are highlighted in bold. QM9 only contains 36 eight-ring molecules and the proportion for eight-ring is nearly 0.}
    \begin{tabular}{c|cccc|ccccc}
    \toprule
    \multicolumn{10}{c}{P (\%)} \\
    \midrule
    Method & 0     & 1     & 2     & 3     & 4     & 5     & 6     & 7     & 8 \\
    \midrule
    QM9   & 10.2  & 39.3  & 27.6  & 15.1  & 4.4   & 2.7   & 0.6   & 0.2   & 0.0 \\
    \midrule
    EDM   & 10.5  & 39.8  & 28.0  & 14.5  & 4.0   & 2.9   & 0.2   & 0.1   & 0.0 \\
    GeoLDM & 12.0  & 38.6  & 27.0  & 15.3  & 4.6   & 2.2   & 0.2   & 0.1   & 0.0 \\
    EquiFM & 12.1  & 44.1  & 29.8  & 11.8  & 1.7   & 0.5   & 0.0   & 0.0   & 0.0 \\
    GeoBFN & 2.8   & 41.5  & 32.1  & 15.7  & 4.7   & 2.7   & 0.3   & 0.1   & 0.0 \\
    \midrule
    C-EDM & 98.9  & 94.2  & 80.8  & 64.4  & 12.6  & 26.8  & 0.3   & 0.1   & 0.0 \\
    C-GeoLDM & 97.1  & 89.4  & 74.2  & 52.4  & 22.3  & 22.7  & 0.9   & 0.2   & 0.0 \\
    EEGSDE & 98.4  & 92.2  & 77.6  & 58.2  & 14.1  & 17.6  & 0.3   & 0.0   & 0.0 \\
    \midrule
    MOOD  & 80.7  & 87.1  & 86.1  & 73.3  & 34.1  & 32.3  & 10.3  & 0.2   & 0.0 \\
    CGD   & 82.3  & 84.8  & 86.2  & 83.6  & 34.4  & 22.4  & 10.3  & 10.1  & 0.0 \\
    \midrule
    DiffLinker & 99.7  & 99.9  & 99.0  & 91.4  & 84.7  & 75.6  & 74.6  & 73.2  & 69.4 \\
    LinkerNet & 99.8  & 99.6  & 88.8  & 87.2  & 83.2  & 73.7  & 66.1  & 64.7  & 59.2 \\
    \midrule
    \textit{GODD} & \textbf{99.9} & \textbf{99.8} & \textbf{99.1} & \textbf{97.6} & \textbf{92.5} & \textbf{89.7} & \textbf{78.7} & \textbf{88.2} & \textbf{82.1} \\
    \bottomrule
    \end{tabular}%
  \label{tab: full ring results-1}%
\end{table}%

\begin{table}[htbp]
  \centering
  \caption{Results of molecule proportion in terms of molecule validity (V). The \textbf{best} results are highlighted in bold. }
    \begin{tabular}{c|cccc|ccccc|c}
    \toprule
    \multicolumn{11}{c}{Target Valid (\%)} \\
    \midrule
    Method & 0     & 1     & 2     & 3     & 4     & 5     & 6     & 7     & 8     & Averaged \\
    \midrule
    EDM   & 10.8  & 36.1  & 26.7  & 13.9  & 4.0   & 2.3   & 0.2   & 0.1   & 0.0   & 10.5 \\
    GeoLDM & 11.2  & 36.2  & 25.2  & 14.3  & 4.3   & 2.0   & 0.2   & 0.1   & 0.0   & 10.4 \\
    EquiFM & 11.4  & 41.4  & 28.0  & 11.1  & 1.6   & 0.5   & 0.0   & 0.0   & 0.0   & 10.4 \\
    GeoBFN & 2.7   & 38.8  & 30.0  & 14.7  & 4.4   & 2.6   & 0.3   & 0.1   & 0.0   & 10.4 \\
    \midrule
    C-EDM & 86.6  & 85.4  & 74.9  & 59.8  & 12.1  & 23.3  & 0.2   & 0.1   & 0.0   & 38.0 \\
    C-GeoLDM & 86.2  & 79.6  & 65.8  & 48.1  & 20.4  & 20.7  & 0.9   & 0.2   & 0.0   & 35.7 \\
    EEGSDE & 96.7  & 92.1  & 77.2  & 58.0  & 13.9  & 17.4  & 0.3   & 0.0   & 0.0   & 39.5 \\
    \midrule
    MOOD  & 75.5  & 81.7  & 80.6  & 68.9  & 32.0  & 30.1  & 9.6   & 0.1   & 0.0   & 42.1 \\
    CGD   & 77.0  & 79.6  & 81.1  & 78.4  & 32.3  & 20.9  & 9.5   & 9.5   & 0.0   & 43.2 \\
    \midrule
    DiffLinker & 64.3  & 88.8  & 92.6  & 86.7  & 80.1  & 71.7  & 62.7  & 48.2  & 37.7  & 70.3 \\
    LinkerNet & 63.9  & 90.2  & 83.4  & 84.4  & 79.9  & 69.8  & 60.9  & 62.0  & 55.0  & 72.2 \\
    \midrule
    \textit{GODD} & \textbf{31.7} & \textbf{91.4} & \textbf{91.4} & \textbf{92.1} & \textbf{85.3} & \textbf{85.2} & \textbf{69.5} & \textbf{82.5} & \textbf{72.2} & \textbf{77.9} \\
    \bottomrule
    \end{tabular}%
  \label{tab: full ring results-2}%
\end{table}%

\begin{table}[htbp]
  \centering
  \caption{Results of molecule proportion in terms of novelty (N). The \textbf{best} results are highlighted in bold.}
    \begin{tabular}{c|cccc|ccccc|c}
    \toprule
    \multicolumn{11}{c}{Target Novelty (\%)} \\
    \midrule
    Method & 0     & 1     & 2     & 3     & 4     & 5     & 6     & 7     & 8     & Averaged \\
    \midrule
    EDM   & 7.1   & 23.6  & 17.5  & 9.1   & 2.6   & 1.5   & 0.1   & 0.1   & 0.0   & 6.8 \\
    GeoLDM & 7.0   & 22.4  & 15.6  & 8.9   & 2.7   & 1.3   & 0.1   & 0.0   & 0.0   & 6.4 \\
    EquiFM & 7.5   & 27.1  & 18.3  & 7.2   & 1.1   & 0.3   & 0.0   & 0.0   & 0.0   & 6.8 \\
    GeoBFN & 1.7   & 25.0  & 19.4  & 9.5   & 2.8   & 1.7   & 0.2   & 0.1   & 0.0   & 6.7 \\
    \midrule
    C-EDM & 57.1  & 59.7  & 54.2  & 44.2  & 9.9   & 20.1  & 0.2   & 0.1   & 0.0   & 27.3 \\
    C-GeoLDM & 63.3  & 61.6  & 53.3  & 40.1  & 17.3  & 18.3  & 0.7   & 0.1   & 0.0   & 28.3 \\
    EEGSDE & 63.9  & 61.4  & 53.0  & 42.5  & 9.9   & 14.1  & 0.3   & 0.0   & 0.0   & 27.2 \\
    \midrule
    MOOD  & 50.0  & 53.9  & 53.6  & 44.4  & 20.6  & 20.0  & 6.3   & 0.1   & 0.0   & 27.6 \\
    CGD   & 51.0  & 52.5  & 53.1  & 51.3  & 21.0  & 13.9  & 6.3   & 6.2   & 0.0   & 28.4 \\
    \midrule
    DiffLinker & 84.8  & 34.0  & 33.7  & 31.1  & 28.8  & 25.7  & 74.5  & 63.2  & 59.4  & 48.4 \\
    LinkerNet & 86.5  & 55.0  & 52.9  & 52.3  & 58.3  & 51.6  & 46.2  & 45.3  & 41.5  & 54.4 \\
    \midrule
    \textit{GODD} & \textbf{96.6} & \textbf{51.3} & \textbf{55.6} & \textbf{60.2} & \textbf{69.5} & \textbf{63.5} & \textbf{71.5} & \textbf{83.4} & \textbf{80.9} & \textbf{70.3} \\
    \bottomrule
    \end{tabular}%
  \label{tab: full ring results-3}%
\end{table}%

\begin{table}[htbp]
  \centering
  \caption{Results of molecule proportion in terms of success rate (S). The \textbf{best} results are highlighted in bold.}
    \begin{tabular}{c|cccc|ccccc|c}
    \toprule
    \multicolumn{11}{c}{Success Rate (\%)} \\
    \midrule
          & 0     & 1     & 2     & 3     & 4     & 5     & 6     & 7     & 8     & Averaged \\
    \midrule
    EDM   & 6.5   & 21.9  & 16.2  & 8.4   & 2.4   & 1.4   & 0.1   & 0.1   & 0.0   & 6.3 \\
    GeoLDM & 6.4   & 20.6  & 14.4  & 8.2   & 2.4   & 1.2   & 0.1   & 0.0   & 0.0   & 5.9 \\
    EquiFM & 6.9   & 25.1  & 17.0  & 6.7   & 1.0   & 0.3   & 0.0   & 0.0   & 0.0   & 6.3 \\
    GeoBFN & 1.6   & 23.0  & 17.8  & 8.7   & 2.6   & 1.5   & 0.2   & 0.1   & 0.0   & 6.1 \\
    \midrule
    C-EDM & 48.1  & 53.8  & 50.0  & 40.5  & 7.9   & 16.8  & 0.2   & 0.1   & 0.0   & 24.1 \\
    C-GeoLDM & 54.6  & 54.6  & 46.9  & 36.8  & 15.4  & 15.6  & 0.6   & 0.1   & 0.0   & 25.0 \\
    EEGSDE & 54.7  & 54.7  & 46.9  & 39.5  & 9.5   & 12.2  & 0.2   & 0.0   & 0.0   & 24.2 \\
    \midrule
    MOOD  & 45.9  & 49.8  & 49.4  & 41.0  & 18.9  & 18.3  & 5.8   & 0.1   & 0.0   & 25.5 \\
    CGD   & 46.8  & 48.5  & 49.1  & 47.3  & 19.5  & 12.8  & 5.8   & 5.7   & 0.0   & 26.2 \\
    \midrule
    DiffLinker & 54.6  & 18.1  & 18.9  & 17.7  & 16.3  & 14.6  & 37.6  & 37.1  & 22.8  & 26.4 \\
    LinkerNet & 55.1  & 49.5  & 34.8  & 35.4  & 39.2  & 34.2  & 29.8  & 32.6  & 22.1  & 37.0 \\
    \midrule
    \textit{GODD} & \textbf{25.9} & \textbf{43.4} & \textbf{46.2} & \textbf{50.4} & \textbf{53.8} & \textbf{41.0} & \textbf{46.1} & \textbf{34.1} & \textbf{23.9} & \textbf{40.5} \\
    \bottomrule
    \end{tabular}%
  \label{tab: full ring results-4}%
\end{table}%

\clearpage

\section{Visualization}
\label{appendix: visualization}
\begin{figure*}[ht]
\begin{center}
\centerline{
    \includegraphics[width=\textwidth]{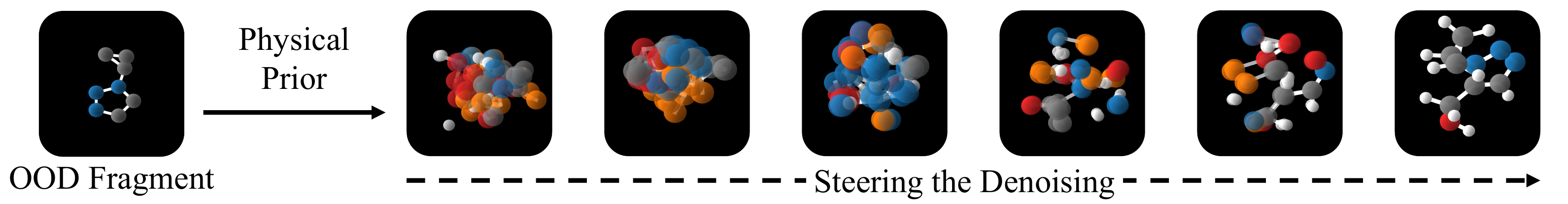}
}
\caption{{\em The Illustration of Generating OOD Samples with~\textit{GODD}}: given an OOD fragment as the structural prior, our trained~\textit{GODD} can generate valid, unique, and novel molecules containing the target fragment.}
\vspace{-2.5em}
\label{fig: scaffold conditional design}
\end{center}
\end{figure*}

\begin{figure}[ht]
\begin{center}
\centerline{
    \includegraphics[width=\columnwidth]{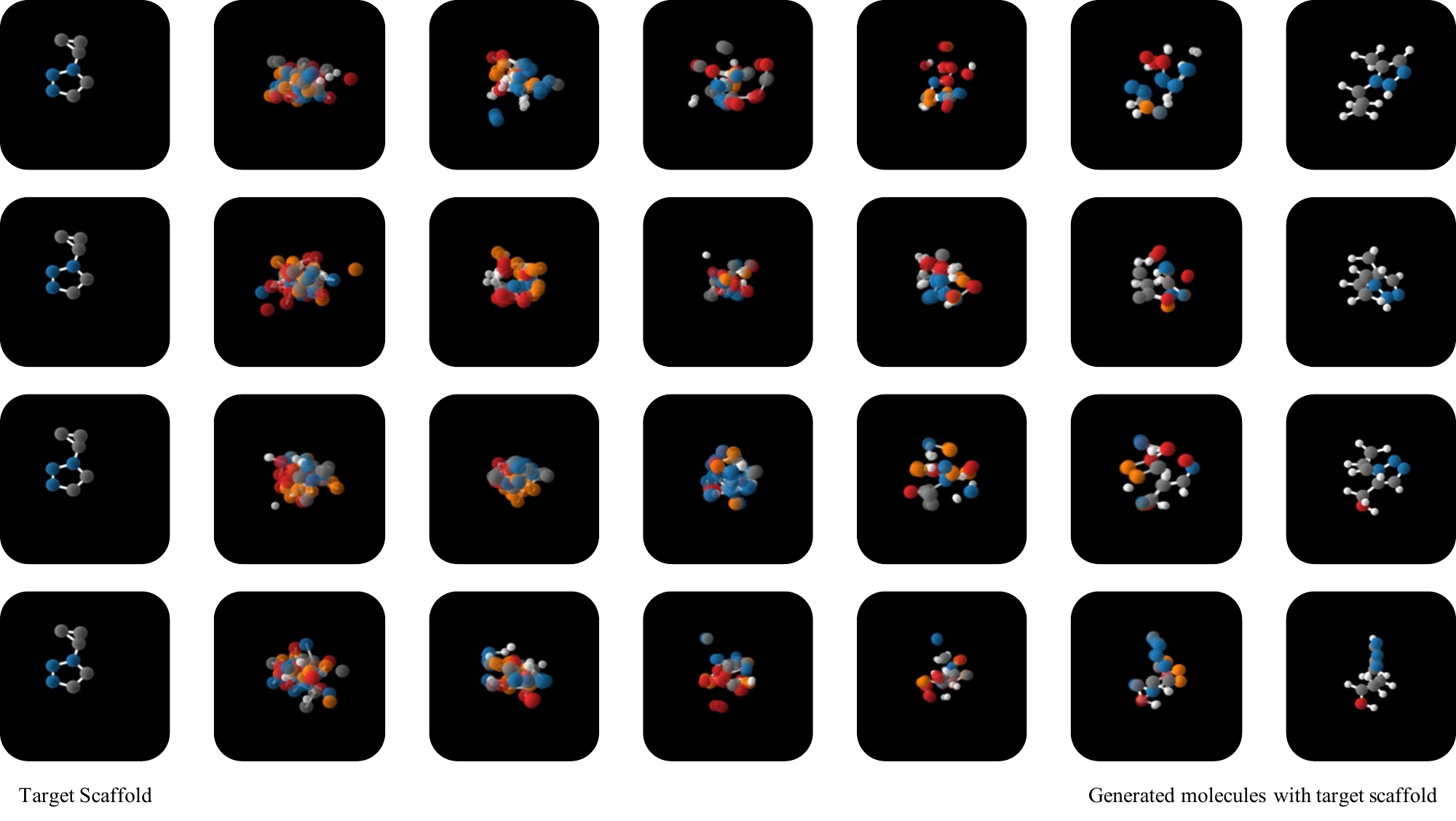}
}

\caption{Molecules Generated by \textit{GODD} for Scaffold Adaptive Generation Under The Same Unseen Scaffold Condition.}
\label{fig: scaffold qm9 chain examples}
\end{center}
\end{figure}

\begin{figure}[ht]
\begin{center}
\centerline{
    \includegraphics[width=\columnwidth]{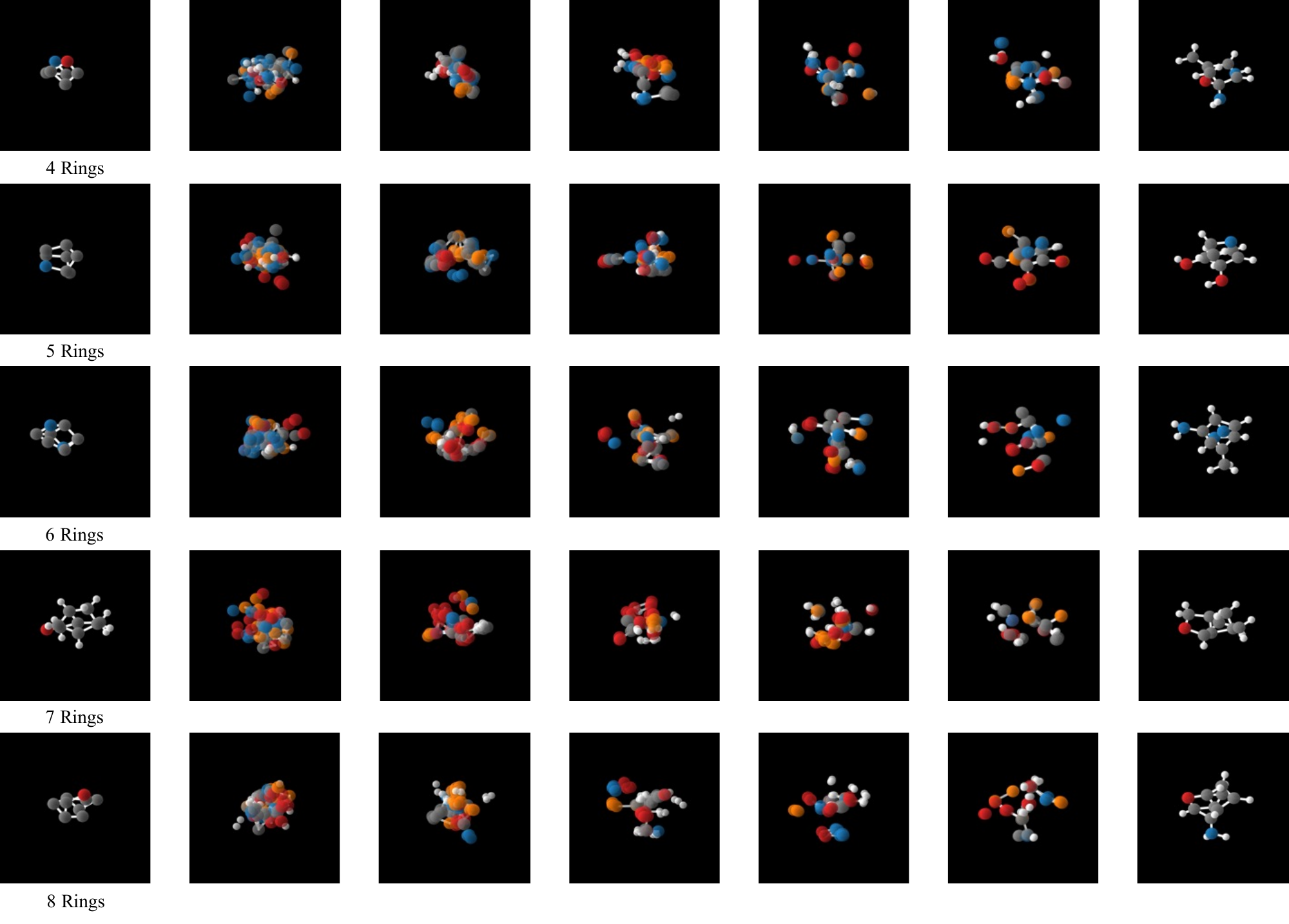}
}
\caption{Molecules Generated by \textit{GODD} for Ring Number Adaptive Generation For Unseen Ring Numbers}
\label{fig: ring qm9 examples}
\end{center}
\end{figure}

\begin{figure}[ht]
\begin{center}
\centerline{
    \includegraphics[width=\columnwidth]{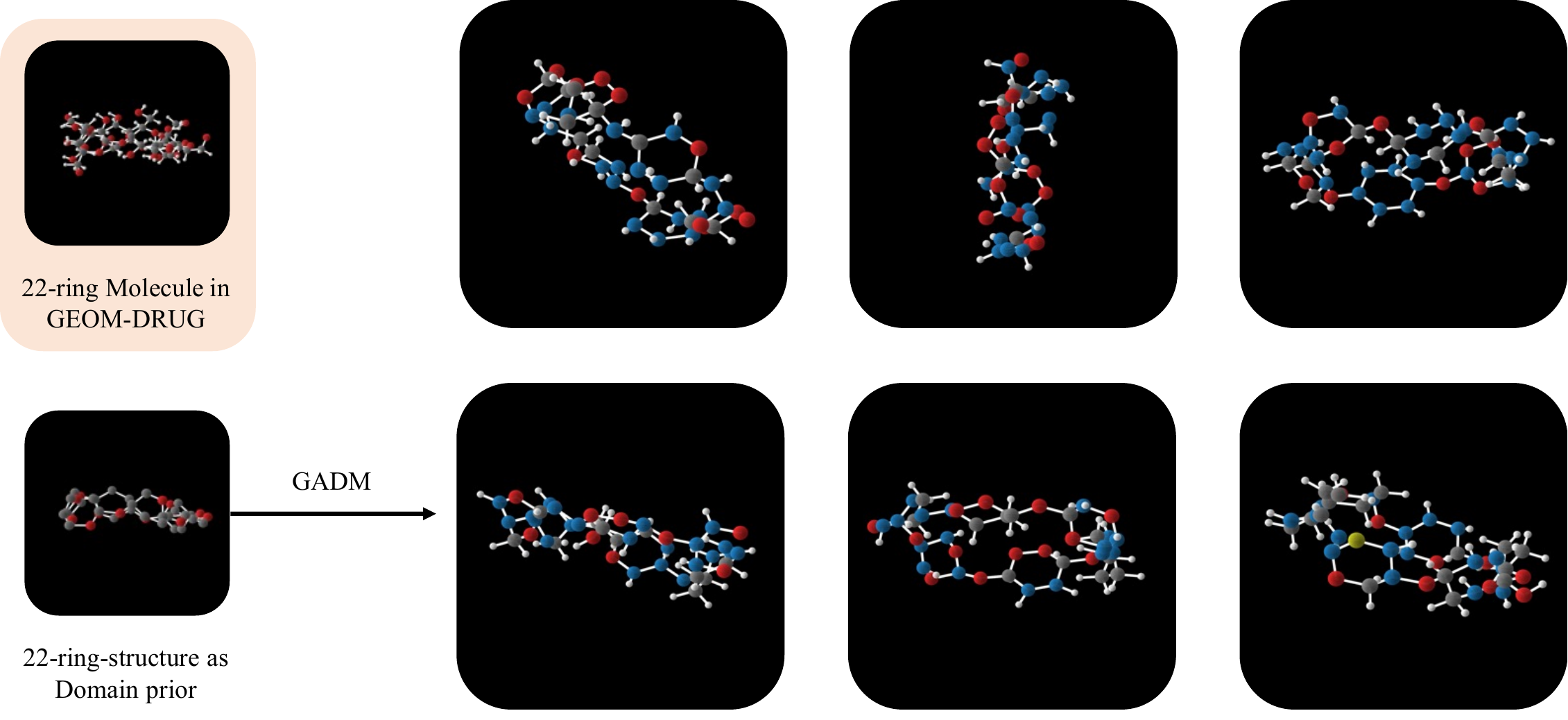}
}
\caption{Molecules Generated by \textit{GODD} for Ring Number Adaptive Generation For Unseen Ring Numbers on GEOM-DRUG Dataset.}
\label{fig: ring drug examples}
\end{center}
\end{figure}

In this section, we provide additional visualizations of structural prior steered molecule generation by \textit{GODD} for OOD scaffold generation and ring number generation in Figures \ref{fig: scaffold conditional design}, \ref{fig: scaffold qm9 chain examples}, \ref{fig: ring qm9 examples}, and \ref{fig: ring drug examples}.

As depicted in the two figures, the model consistently generates realistic molecular geometries with OOD scaffolds or ring numbers.

\clearpage

\section{Related Work}
\label{sec: related work}
\textbf{Molecule Generation Models.}
Prior studies on molecule generation focused on generating molecules as 2D graphs~\cite{pmlr-v80-jin18a, NEURIPS2018_b8a03c5c, Shi*2020GraphAF:}. However, there has been a growing interest in 3D molecule generation. G-SchNet~\cite{NEURIPS2019_a4d8e2a7} and G-SphereNet~\cite{luo2022an} utilize autoregressive techniques to construct molecules incrementally by progressively connecting atoms or molecular fragments. These frameworks necessitate either a meticulous formulation of a complex action space or action ordering.
\par
More recently, the focus has shifted towards using diffusion models and flow matching models for 3D molecule generation~\cite{pmlr-v162-hoogeboom22a, pmlr-v202-xu23n,wu2022diffusionbased, Song2024unified, hong2025accelerating}. To mitigate the inconsistency of unified Gaussian diffusion across diverse modalities, a latent space was introduced by~\cite{pmlr-v202-xu23n}. To tackle the atom-bond inconsistency problem, different noise schedulers were proposed by~\cite{peng2023moldiff} for various modalities to accommodate noise sensitivity. However, these algorithms do not account for generating novel molecules outside the training distribution.

\textbf{Out-of-Distribution Molecule Generation.} OOD generation, although under-explored, is of paramount importance, especially considering that molecules generated by machine-learning methods often exhibit a ``striking similarity"~\cite{RN364}. 
In recent years, some preliminary work has begun to use reinforcement learning~\cite{NEURIPS2021_41da609c} and out-of-distribution control~\cite{exploring-ood} to explore the generation of novel molecules. However, these methods are still challenging when designing novel molecules in data-sparse regions with fragment shifts. As proposed by~\cite{exploring-ood}, MOOD employs an OOD control and integrates a property-predictor-based diffusion scheme to optimize molecules for specific chemical properties. Similarly, CGD~\cite{context-ood} leverages unlabeled data to improve the generalization of guided diffusion models. However, these predictor-based OOD methods fail to generate novel molecules with ODD fragments that are sparse for training a classifier.
\par
\textbf{Fragment-Based Drug Design.} 
The discovery of new molecules is crucial across various fields, and there are four primary approaches to this task~\cite{fbdd-rise}: (1) searching from an existing molecule, (2) developing from a natural product, (3) high-throughput screening, and (4) fragment-based drug discovery (FBDD). Among these, FBDD has gained significant importance and interest over the past decades due to its higher efficiency compared to other methods~\cite{fbdd-rise}. Typically, fragments are selected based on the ``rule of three''~\cite{rule-3-for-fragment} criteria and thereby can be grown, linked, or merged to develop potential molecules~\cite{computational-fbdd}. Recently, there has been a growing trend in enhancing FBDD with machine learning techniques~\cite{smiles-fbdd, linker-design-NMI, linkernet-nips}. However, these methods often overlook the issue of fragment sparsity in datasets, highlighting the need for an OOD molecular generative model capable of producing realistic molecules in data-sparse regions.

\section{Limitations}
\label{appendix: limitation}
This paper addresses the problem of OOD generation in the context of structural shifts. However, in some scenarios, OOD structures may not be provided. We plan to investigate this issue in future work by developing methods to identify structural variations when OOD structures are unavailable. Additionally, most generative models, including ours, adopt the EGNN modules to capture the equivariance of molecules~\cite{pmlr-v162-hoogeboom22a, pmlr-v202-xu23n, equifm, geobfn}. The model’s memory overhead escalates exponentially with the size of the input molecules, posing a significant constraint, especially for generating large molecules. Given a molecule $\mathcal{G}=\langle\mathbf{x}\in\mathbb{R}^{n\times3},\mathbf{h}\in\mathbb{R}^{n\times f}\rangle$. Suppose the total number of layers of EGNNs used is $l$ and the hidden feature for EGNN is $h$, then the space complexity of our model is $\mathcal{O}(nnhl)$. For example, in the GEOM-DRUG dataset, if molecules of 180 atoms are processed, all methods EGNN-based algorithms require around 3.5GB of memory, which results in huge overhead for experiments.

\section{Impact Statements}
\label{appendix: impact}
This paper presents work whose goal is to advance the field of generative Artificial Intelligence (AI) for scientific fields, such as material science, chemistry, and biology. The obtained experience/knowledge will greatly boost generative AI technologies in facilitating the process of scientific knowledge discovery.
\par
Machine learning for molecule generation opens up possibilities for designing molecules beyond therapeutic purposes, such as the creation of illicit drugs or dangerous substances. The potential for misuse and unintended consequences necessitates strict ethical guidelines, robust regulation, and responsible use of these technologies to prevent harm to individuals and society.


\end{document}